\documentclass{article}

\PassOptionsToPackage{backref}{hyperref}

\usepackage[final, nonatbib]{neurips_2024}                 
\usepackage[utf8]{inputenc}                      
\usepackage[T1]{fontenc}                         
\usepackage{url}                                 
\usepackage{booktabs}                            
\usepackage{amsfonts}                            
\usepackage{nicefrac}                            
\usepackage{microtype}                           
\usepackage{xcolor}                              

\usepackage[pagebackref]{hyperref}               
\usepackage{amsmath}                             
\usepackage{amssymb}                             
\usepackage[pdftex]{graphicx}                    
\usepackage[noabbrev,capitalise]{cleveref}       
\usepackage{multirow}                            
\usepackage{algorithm}                           
\usepackage{algorithmicx}                        
\usepackage{algpseudocode}                       
\usepackage{enumitem}                            
\usepackage{lipsum}
\usepackage{caption}                             
\usepackage{subcaption}                          
\usepackage{makecell}                            
\usepackage{pdflscape}
\usepackage{pgfplots}
\usepackage{listings}
\usepackage{pifont}
\usepackage{cancel}
\usepackage{diagbox}
\usepackage{siunitx}
\usepackage{tikz}
\usetikzlibrary{plotmarks}

\definecolor{wikiblue}{HTML}{3366CC}
\hypersetup{colorlinks=true, linkcolor=wikiblue, citecolor=wikiblue, urlcolor=wikiblue}
\input{commands.tex}

\title{Predictive Attractor Models}

\author{%
  Ramy Mounir\qquad
  Sudeep Sarkar \\
  Department of Computer Science and Engineering,
  University of South Florida, Tampa\\
  \texttt{\{ramy, sarkar\}@usf.edu}
}

\begin{document}
  \maketitle

  \begin{abstract}
    Sequential memory, the ability to form and accurately recall a sequence of events or stimuli in the correct order, is a fundamental prerequisite for biological and artificial intelligence as it underpins numerous cognitive functions (e.g., language comprehension, planning, episodic memory formation, etc.) However, existing methods of sequential memory suffer from catastrophic forgetting, limited capacity, slow iterative learning procedures, low-order Markov memory, and, most importantly, the inability to represent and generate multiple valid future possibilities stemming from the same context. Inspired by biologically plausible neuroscience theories of cognition, we propose \textit{Predictive Attractor Models (PAM)}, a novel sequence memory architecture with desirable generative properties. PAM is a streaming model that learns a sequence in an online, continuous manner by observing each input \textit{only once}. Additionally, we find that PAM avoids catastrophic forgetting by uniquely representing past context through lateral inhibition in cortical minicolumns, which prevents new memories from overwriting previously learned knowledge. PAM generates future predictions by sampling from a union set of predicted possibilities; this generative ability is realized through an attractor model trained alongside the predictor. We show that PAM is trained with local computations through Hebbian plasticity rules in a biologically plausible framework. Other desirable traits (e.g., noise tolerance, CPU-based learning, capacity scaling) are discussed throughout the paper. Our findings suggest that PAM represents a significant step forward in the pursuit of biologically plausible and computationally efficient sequential memory models, with broad implications for cognitive science and artificial intelligence research.
\end{abstract}

  \section{Introduction}
\label{sec:introduction}

Modeling the temporal associations between consecutive inputs in a sequence (i.e., sequential memory) enables biological agents to perform various cognitive functions, such as episodic memory formation~\cite{rolls2010computational, tulving1972episodic, conway2009episodic}, complex action planning~\cite{haggard1998planning} and translating between languages~\cite{afraimovich2015sequential}. For example, playing a musical instrument requires remembering the sequence of notes in a piece of music; similarly, playing a game of chess requires simulating, planning, and executing a sequence of moves in a specific order. While the ability to form such memories of static, unrelated events has been extensively studied~\cite{salvatori2021associative, salvatori2022learning, tang2023recurrent, yoo2022bayespcn, whittington2017approximation}, the ability of biologically-plausible artificial networks to learn and recall temporally-dependent patterns has not been sufficiently explored in literature~\cite{tang2024sequential}. The task of sequential memory is considered challenging for models operating under biological constraints (i.e., local synaptic computations) for many reasons, including catastrophic forgetting, ambiguous context representations, multiple future possibilities, etc. 

In addition to the biological constraint, we impose the following set of desirable characteristics as learning constraints on sequence memory models.

\begin{itemize}
    \item The learning of one sequence does not overwrite the previously learned sequences. This property is defined under the continual learning framework as avoiding catastrophic forgetting and evaluated with the Backward Transfer (BWT) metric~\cite{lopez2017gradient}.
    \item The model should uniquely encode inputs based on their context in a sequence. Consider the sequence of letters ``EVER''; the representation of ``E'' at position 1 should be different from ``E'' at position 3, thus resulting in different predictions: ``V'' and ``R''. Moreover, the representation of ``E'' at position 3 in ``EVER'' should be different from ``E'' at position 3 in ``CLEVER''. Therefore, positional encoding is not a valid solution.
    \item When presented with multiple valid, future possibilities, the model should learn to represent each possibility separately yet stochastically sample a single valid possibility. Consider the two sequences ``THAT'' and ``THEY''; after seeing ``TH'', the model should learn to generate either ``A'' or ``E'', but not an average~\cite{lecun2022path} or a union of both~\cite{hawkins2016neurons}.
    \item The model should be capable of incrementally learning each transition without seeing the whole sequence or revisiting older sequence transitions that are previously learned. This property falls under online learning constraints, also called stream learning~\cite{mounir2024streamer, hawkins2016neurons}.
    \item The learning algorithm should be tolerant and robust to significant input noise. A model should continuously clean the noisy inputs using learned priors and beliefs, thus performing future predictions based on the noise-free observations~\cite{hawkins2016neurons}.
\end{itemize}

We propose \textit{Predictive Attractor Models (PAM)}, which consists of a state prediction model and a generative attractor model. The predictor in PAM is inspired by the Hierarchical Temporal Memory (HTM)~\cite{hawkins2016neurons} learning algorithm, where a group of neurons in the same cortical minicolumn share the same receptive feedforward connection from the sensory input on their proximal dendrites. The depolarization of the voltage of any neuron in a single minicolumn (i.e., on distal dendrites) primes this depolarized neuron to fire first while inhibiting all the other neurons in the same minicolumn from firing (i.e., competitive learning). The choice of which neurons fire within the same minicolumn is based on the previously active neurons and their trainable synaptic weights to the depolarized neurons, which gives rise to a unique context representation for every input. The sparsity of representations (discussed later in Section~\ref{sec:prelim}) allows for multiple possible predictions to be represented as a union of individual cell assemblies. The Attractor Model learns to disentangle possibilities by strengthening the synaptic weights between active neurons of input representations and inhibiting the other predicted possibilities from firing, effectively forming fixed point attractors during online learning. During recall, the model uses these learned conditional attractors to sample one of the valid predicted possibilities or uses the attractors as prior beliefs for removing noise from sensory observations.

PAM satisfies the above-listed constraints for a sequential memory model, whereas the current state-of-the-art models fail in all or many of the constraints, as shown in the experiments. Our contributions can be summarized as follows: 
(1) Present the novel \textit{PAM} learning algorithm that can explicitly represent context in memory without backpropagation, avoid catastrophic forgetting, and perform stochastic generation of multiple future possibilities. 
(2) Perform extensive evaluation of PAM on multiple tasks (e.g., sequence capacity, sequence generation, catastrophic forgetting, noise robustness, etc.) and different data types (e.g., protein sequences, text, vision).
(3) Formulate PAM and its learning rules as a State Space Model grounded in variational inference and the Free Energy Principle~\cite{friston2010free}.

  \section{Background and Related Works}\label{sec:background}

\textbf{Predictive Coding}\; Predictive coding proposes a framework for the hierarchical processing of information. It was initially formulated as a time series compression algorithm to create a more efficient coding system~\cite{elias1955predictive,o1971entropy}. A few decades later, PC was used to model visual processing in the Retina~\cite{srinivasan1982predictive,hosoya2005dynamic} as an inference model. In the seminal work of Rao and Ballard~\cite{rao1999predictive}, PC was reformulated as a general computational model of the cortex. The main intuition is that the brain continuously predicts all perceptual inputs, resulting in a quantity of prediction error which can be minimized by adjusting its neural activities and synaptic strengths. In-depth variational free energy derivation is provided in Appendix~\ref{app:VFE}.

PC defines two subgroups of neurons: value $\boldsymbol{z}$ and error. Each neuron contains a value node sending its prediction to the lower level $\hat{\boldsymbol{z}}_l = f_{l+1}(\boldsymbol{z}_{l+1})$ through learnable function $f$, and error node propagating its computed error to the higher level. The total prediction error is computed as $\epsilon =  \sum_{l} ||(\boldsymbol{z}_l - \hat{\boldsymbol{z}}_l)||_2^2$, which is minimized by first running the network value nodes to equilibrium through optimizing the value nodes $\{\boldsymbol{z_l}\}_{l=0}^L$. At equilibrium, the value nodes are fixed, and inferential optimization is performed by optimizing the functions $\{f_l\}_{l=1}^L$. Both optimizations aim to minimize the same prediction error over all neurons. This propagation of error to equilibrium is shown to be equivalent to backpropagation but using only local computations~\cite{whittington2017approximation}. The PC formulation has shown success in training on static and i.i.d data~\cite{whittington2017approximation, yoo2022bayespcn, salvatori2022incremental, han2018deep}. More recently~\cite{tang2024sequential}, Temporal Predictive Coding (tPC) has also shown some success in sequential memory tasks by modifying error formulation to account for a one-step synaptic delay through interneurons, thus modeling temporal associations between sequence inputs. In the experiments, we compare our model to tPC and its 2-layer variant~\cite{tang2024sequential}. Other PC-inspired models, such as~\cite{lotter2016deep, han2018deep, han2019video}, have diverged from the biologically plausible constraints by training through backpropagation through time.

\textbf{Fixed-Point Attractor Dynamics}\; Attractor dynamics refer to mathematical models that describe the behavior of dynamical systems. In our review, we focus on fixed point attractors, specifically Hopfield Networks~\cite{hopfield1982neural} (HN), which is an instance of associative memory models~\cite{kosko1988bidirectional, kage2023implementing, kanerva1988sparse}. Consider, an ordered sequence of $T+1$ consecutive patterns  $ \boldsymbol{x} = [\boldsymbol{x_t}]_{t=1}^{T+1}$, where $\boldsymbol{x}_t \in \{-1,1\}^{N}$. We refer to the Universal Hopfield Networks (UHN) framework~\cite{millidge2022universal} to describe all variants of HN architecture using a similarity (sim) function and a separation (sep) function, as shown in equation~\ref{eqn:UHN-cases}. This family of models can be viewed as a memory recall function, where a query $\xi$ (i.e., noisy or incomplete pattern) is compared to the existing patterns to compute similarity scores using the ``sim'' function. These scores are then used to weight the projection patterns after applying a ``sep'' function to increase the separation between similarity scores. The classical HN uses symmetric weights to store the patterns; therefore, it cannot be used to model temporal associations in a sequence. The asymmetric Hopfield Network (AHN)~\cite{sompolinsky1986temporal} uses asymmetric weights to recall the next pattern in the sequence for a given query $\xi$.
\begin{equation}\label{eqn:UHN-cases}
    \hat{\xi} = \underbrace{P}_{\text{Projection}} \cdot \underbrace{\text{sep}}_{\text{Separation}}(\underbrace{\text{sim}(M,\xi)}_{\text{Similarity}}) =
    \begin{cases}
        \sum_{t=1}^T \mathbf{x}_{t+1}\text{sep}\left(\text{sim}(\mathbf{x}_{t}, \xi)\right) \quad \text{Asymmetric Weights}\\
        \sum_{t=1}^{T+1} \mathbf{x}_t\text{sep}\left(\text{sim}(\mathbf{x}_{t}, \xi)\right) \quad\quad \text{Symmetric Weights}
    \end{cases}
\end{equation}

When a dot product ``sim'' function and identity ``sep'' function are used, we get the classical HN~\cite{hopfield1982neural} and AHN~\cite{sompolinsky1986temporal}. A few variants have been proposed to increase the capacity of the model. Recently, \cite{chaudhry2024long} has extended AHN by using a polynomial (with degree $d$) or a softmax function (with temperature $\beta$) as the ``sep'' function. HN can also be applied to continuous dense patterns~\cite{krotov2016dense, ramsauer2020hopfield, chaudhry2024long}.

\textbf{Predictive Learning}\; Predictive learning takes a more general form of minimizing the prediction error between two views of the same input to improve representations. Many backpropagation-based approaches to predictive learning have been proposed; most recently, JEPA~\cite{lecun2022path} and its variants~\cite{assran2023self, bardes2023mc, bardes2024revisiting, zbontar2021barlow}, learn useful dense representations from images and videos using the predictive objective. Other models, such as~\cite{caron2021emerging, grill2020bootstrap, chen2021exploring} - to list a few, use similar methodology of predicting distorted versions of the same input to learn good feature representations. Prediction-based approaches have also been used to segment videos into events temporally~\cite{aakur2019perceptual, mounir2022self} and spatially~\cite{mounir2023towards, mounir2021spatio, aakur2020action}. More recently, STREAMER~\cite{mounir2024streamer} used a predictive learning approach to achieve hierarchical segmentation and representation learning from streaming egocentric videos, where a high prediction error is used as an event boundary. While these biologically implausible approaches show impressive results on their respective tasks, they still suffer from deep learning known limitations, such as catastrophic forgetting and the inability to generate multiple possibilities in regression-based predictive tasks. \textit{Hierarchical Temporal Memory} (HTM)~\cite{hawkins2016neurons} is a predictive approach that is heavily inspired by neurobiology. HTM relies on lateral inhibition between neurons of the same minicolumn and sparsity of input representations (i.e., SDR) to learn temporal context and associations using only local Hebbian rules. HTM can be applied to online tasks, such as anomaly detection~\cite{ahmad2017unsupervised}, but it is currently incapable of generating future predictions in auto-regressive prediction tasks.

  \section{Predictive Attractor Models}
\label{sec:pam}

\subsection{State Space Model (SSM) formulation}

\begin{figure}[h]
\centering
\includegraphics[width = \linewidth]{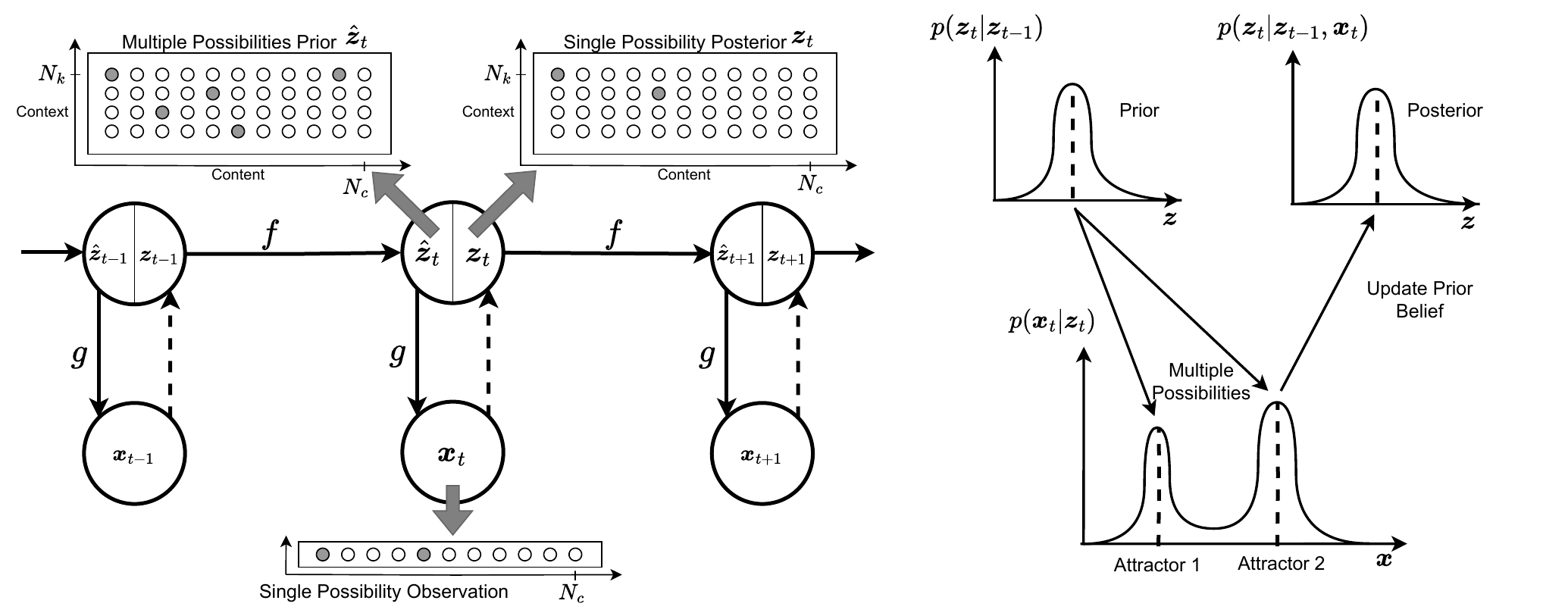}
\caption{State Space Model. (Left): Dynamical system represented by first-order Markov chain of latent states $\boldsymbol{z}$ with transition function $f$ and an emission function $g$ which projects to the observation states $\boldsymbol{x}$. (Right): Gaussian form assumptions for the prior $\hat{\boldsymbol{z}}$ and posterior $\boldsymbol{z}$ latent states, and the Mixture of Gaussian model representing the conditional probability of multiple possibilities $p(\boldsymbol{x}|\boldsymbol{z})$} \label{fig:ssm}
\end{figure}

PAM can be represented as a dynamical system with its structure depicted by a Bayesian probabilistic graphical~\cite{lauritzen1996graphical, koller2009probabilistic} model, more specifically, a State Space Model, where we can perform Bayesian inference on the latent variables and derive learning rules using Variational Inference. Formally, we define a state space model as a tuple $(\mathcal{Z}, \mathcal{X}, f, g)$, where $\mathcal{Z}$ is the latent state space, $\mathcal{X}$ is the observation space, and $f$ and $g$ are the transition and emission functions respectively (similar to HMM~\cite{rabiner1986introduction}). We consider a Gaussian form with white Gaussian noise covariance $\boldsymbol{\Sigma}_z$ for the latent states. However, we assume a latent state $\boldsymbol{z}$ can generate multiple valid possibilities. Therefore, we model the conditional probability $p(\boldsymbol{x}_t | \boldsymbol{z}_t)$ as a Multivariate Gaussian Mixture Model (GMM), where each mode is considered a possibility or a fixed-point attractor in an associative memory model. The GMM has $C$ components with means $g_{c}(\boldsymbol{z}_{t})$, covariances $\boldsymbol{\Sigma}_{c}$ and component weights $w_c$. The SSM dynamics can be formally represented with the following equations:
\begin{equation}~\label{eqn:pam-ssm}
    \boldsymbol{z}_t | \boldsymbol{z}_{t-1} \sim \mathcal{N}(f(\boldsymbol{z}_{t-1}), \boldsymbol{\Sigma}_z), \quad \text{and} \quad
    \boldsymbol{x}_t | \boldsymbol{z}_{t} \sim \sum_{c=1}^{C} w_c \cdot \mathcal{N}(g_{c}(\boldsymbol{z}_{t}), \boldsymbol{\Sigma}_{c})\,,
\end{equation}

where $\boldsymbol{z}_t \in \mathcal{Z}$ and $\boldsymbol{x}_t \in \mathcal{X}$. From the Bayesian inference viewpoint, we are interested in the posterior $p(\boldsymbol{z}_t | \boldsymbol{x}_t, \boldsymbol{z}_{t-1})$. Since the functions $f$ and $g$ are non-linear, the computation of this posterior is intractable (unlike a LG-SSM, such as Kalman Filter~\cite{welch1995introduction}). Therefore, we utilize variational inference to approximate the posterior by assuming the surrogate posterior $q(\boldsymbol{z})$ has a Gaussian form, and minimize the Variational Free Energy (VFE)~\cite{friston2007variational}. The minimization of VFE (in equation~\ref{eqn:pam-vfe}) minimizes the KL-divergence between the approximate posterior $q(\boldsymbol{z})$ and the true posterior $p(\boldsymbol{z}_t | \boldsymbol{x}_t, \boldsymbol{z}_{t-1})$. Derivation~\ref{der:VFE-PAM} of Variational Free Energy is provided in appendix~\ref{app:VFE}
\begin{equation}\label{eqn:pam-vfe}
    \underbrace{
    \sum_{t=1}^{T}
    \mathbb{E}_q \left[\log \left( \frac{q(\boldsymbol{z}_t)} {p(\boldsymbol{x}_t, \boldsymbol{z}_t | \boldsymbol{z}_{t-1})}\right)\right]}_{\text{Variational Free Energy}}
    =
    \underbrace{
    \sum_{t=1}^{T}
    \mathbb{E}_q \left[\frac{1}{\log ( p(\boldsymbol{z}_t | \boldsymbol{z}_{t-1}))}\right]}_{\text{Latent State Error}}
    +
    \underbrace{
    \sum_{t=1}^{T}
    \mathbb{E}_q \left[\frac{1}{\log ( p(\boldsymbol{x}_t | \boldsymbol{z}_t))}\right]}_{\text{Observation Error}}
    -
    \mathcal{H}_q\,,
\end{equation}

where $\mathbb{E}_q \equiv \mathbb{E}_{\boldsymbol{z}_t \sim q(\boldsymbol{z}_t)}$ and $\mathcal{H}_q$ is the Entropy of the approximate posterior $q(\boldsymbol{z})$. The assumption of Gaussian forms for the latent and observable states can further simplify the negative log-likelihood terms (i.e., Latent State Accuracy and Observation Accuracy) to prediction errors. This learning objective encourages the approximate posterior $q(z)$ to assign a high probability to states that explain the observations well and follow the latent dynamics of the system. We minimize the prediction errors (i.e., learn the transition and emission functions) through Hebbian rules as shown in equations~\ref{eqn:heb_A}~and~\ref{eqn:heb_B}.

\begin{theorem}\label{theorem:gmm}

    \newcommand{\bx}{\boldsymbol{x}}
    \newcommand{\bz}{\boldsymbol{z}}
    \newcommand{\ddx}{\frac{\partial}{\partial \bx}}
    \newcommand{\gm}{\mathcal{N}(\bx; \boldsymbol{\mu}_c, \boldsymbol{\Sigma}_c) }
    \newcommand{\cs}{\sum_{c=1}^{C} w_c }
    \newcommand{\gmm}{ \cs \cdot \gm}
    \newcommand{\dgmm}{ - \boldsymbol{\Sigma}_c^{-1} (\bx-\boldsymbol{\mu}_c) }

    Assume the likelihood $p(\boldsymbol{x}_t|\boldsymbol{z}_t)$ in eqn~\ref{eqn:pam-vfe} represents multiple possibilities using a Gaussian Mixture Model (GMM) conditioned on the latent state $\boldsymbol{z}_t$, as shown in eqn~\ref{eqn:pam-ssm}. The maximization of such log-likelihood function (i.e., $\frac{\partial}{\partial \boldsymbol{x}} \log p(\boldsymbol{x}_t|\boldsymbol{z}_t)$) w.r.t a query observation state $\boldsymbol{x}$ is equivalent to the Hopfield recall function (i.e., eqn~\ref{eqn:UHN-cases}) with the means of the GMM representing the attractors of a Hopfield model. Formally, the weighted average of the GMM means (i.e., $\{\boldsymbol{\mu}_c\}_{c=1}^C$), with the weights being a similarity measure, maximizes the log-likelihood of $\boldsymbol{x}$ under the mixture model.
    \begin{equation}
        \bx = \sum_{c=1}^C  \underbrace{\frac{w_c \cdot \gm \cdot \boldsymbol{\Sigma}_c^{-1}}{\gmm \cdot \boldsymbol{\Sigma}_c^{-1}}}_{\text{similarity score}}
        \cdot
        \underbrace{\boldsymbol{\mu}_c}_{\text{projection}}\\
    \end{equation}
\end{theorem}

\textbf{Proof:}\; See derivation~\ref{der:GMM} in appendix~\ref{app:GMM} for the full proof.

\subsection{Preliminaries and Notations}\label{sec:prelim}

\textbf{Sparse Distributed Representation (SDR)}\; Inspired by the sparse coding principles observed in the brain, SDRs encode information using a small set of active neurons in high dimensional binary representation. We adopt SDRs as a more biologically plausible cell assembly representation~\cite{kanerva1988sparse}. SDRs have desirable characteristics, such as a low chance of false positives and collisions between multiple SDRs and high representational capacity~\cite{ahmad2016neurons} (More on SDRs in Appendix~\ref{app:sdr}). An SDR is parameterized by the total number of neurons $N$ and the number of active neurons $W$. The ratio $S = W/N$ denotes the SDR sparsity. A 1-dimensional SDR $\boldsymbol{x}$ can be indexed as $\boldsymbol{x}^i \in \{0, 1\}$, whereas a 2-dimensional SDR $\boldsymbol{z}$ can be indexed as $\boldsymbol{z}^{ij} \in \{0, 1\}$. To identify the active neurons, we define the function $I: \{0, 1\}^{N} \mapsto \mathbb{N}^{W}$ to represent the indices of the active neurons in an SDR $\boldsymbol{x}$ as $I(\boldsymbol{x}) = \{i | \boldsymbol{x}^{i} = 1 \}$.

\textbf{Context as Orthogonal Dimension}\; We transform the high-order Markov dependencies between observation states into a first-order Markov chain of latent states by storing context information in those latent states. The latent states SDRs, $\boldsymbol{z} \in \{0, 1\}^{N_c \times N_k}$, are represented with two orthogonal dimensions, where content information about the input is stored in one dimension with size $N_c$, while context related information is stored in an orthogonal dimension with size $N_k$. Therefore, the projection of the latent state $\boldsymbol{z}$ on the first dimension (i.e., $\downarrow\boldsymbol{z}$) removes all context information from the state. In contrast, adding context information to an observation state $\boldsymbol{x}$ expands the dimensionality of the state (i.e., $\uparrow\boldsymbol{x}$) such that context can be encoded without affecting its content. Competitive learning is enforced on the context dimension through lateral inhibition, effectively minimizing the collisions between contexts of multiple SDRs. We define a projection operator $\downarrow : \{0, 1\}^{N_c \times N_k} \mapsto \{0, 1\}^{N_c} $. Additionally, we define a projection operator $\uparrow: \{0, 1\}^{N_c} \mapsto \{0, 1\}^{N_c \times N_k}$ for 1-dimensional SDRs (i.e., $\boldsymbol{x}$) as shown in equation~\ref{eqn:sdr_op}.
\begin{equation}\label{eqn:sdr_op}
    (\downarrow \boldsymbol{z})^i =
    \begin{cases} 
    1 & \text{if } \exists j \text{ s.t. } \boldsymbol{z}^{ij} = 1, \\ 
    0 & \text{otherwise,} 
    \end{cases}, 
    \quad
    (\uparrow \boldsymbol{x})^{ij} =
    \begin{cases} 
    1 & \text{if } \boldsymbol{x}^i = 1, \\ 
    0 & \text{otherwise,} 
    \end{cases}
\end{equation}

\subsection{Sequence Learning}

Given a sequence of $T+1$ SDR patterns $[\boldsymbol{x}_t]_{t=1}^{T+1}$, where $\boldsymbol{x}_t \in \{0, 1\}^{N_c}$, the sequence can be learned by modeling the context-dependent transitions between consecutive inputs within the sequence. We define learnable weight parameters for transition and emission functions, $\boldsymbol{A} \in \mathbb{R}^{N_c N_k \times N_c N_k}, \boldsymbol{B} \in \mathbb{R}^{N_c \times N_c}$  respectively. A single latent state transition is defined as $\hat{\boldsymbol{z}}_t = \delta(\boldsymbol{A} \cdot \boldsymbol{z}_{t-1}) = \delta(\boldsymbol{a}_t)$, where $\delta$ is a threshold function transforming the logits $\boldsymbol{a}_t$ to the predicted SDR state $\hat{\boldsymbol{z}}_t$. The full sequence learning algorithm is provided in algorithm~\ref{alg:learn}.

\textbf{Context Encoding through Competitive Learning}\; Every observation $\boldsymbol{x}_t$ contains only content information about the input; we embed the observation with context by expanding the state with an orthogonal dimension (i.e., $\uparrow \boldsymbol{x}_t$) which activates all neurons in the minicolumns at the indices $I(\boldsymbol{x}_t)$. Then, for each active minicolumn, the neuron in a predictive state (i.e., higher than the prediction threshold) fires and inhibits all the other neurons in the same minicolumn from firing (i.e., lateral inhibition), as shown in Equation~\ref{eqn:context_embedding}. If none - or more than one - of the neurons are in a predictive state, random Gaussian noise ($\epsilon$) acts as a tiebreaker to select a context neuron. We \textit{do not} allow multiple neurons to fire in the same minicolumn, which is different from HTM~\cite{hawkins2016neurons}, where multiple cells can fire in any minicolumn (e.g., bursting).
\begin{equation}\label{eqn:context_embedding}
    m(\boldsymbol{a}_t)^{ij} =
    \begin{cases} 
    1 & \text{if } \boldsymbol{a}_t^{ij} = \max(\{ \delta(\boldsymbol{a}_t^{ij})+\epsilon \}_{j=0}^{N_k}), \\ 
    0 & \text{otherwise,} 
    \end{cases}, 
    \quad
    \boldsymbol{z}_t = (\uparrow \boldsymbol{x}_t) \cap m(\boldsymbol{a}_t)
\end{equation}

\algrenewcommand\algorithmicindent{0.5em}%
\begin{figure}[t]
\newcommand{\CComment}[1]{\textcolor{gray!80}{\Comment{#1}}}
\begin{minipage}[t]{0.485\textwidth}
\begin{algorithm}[H]
  \small
    \caption{
        : \textbf{Sequence Learning}. Given a sequence $\boldsymbol{x}$ of T+1 patterns, this algorithm learns the Transition and Emission synaptic weights ($\boldsymbol{A}$ and $\boldsymbol{B}$). Fixed start context $m(\boldsymbol{a}_0)$ is initialized for all learned sequences.
    }
    \label{alg:learn}
    \begin{algorithmic}[1]
    \Procedure{Train}{$\boldsymbol{x}$}
        \State $\boldsymbol{z}_0 = (\uparrow \boldsymbol{x}_0) \cap m(\boldsymbol{a}_0)$ \CComment{Eqn.~\ref{eqn:context_embedding}}
        \For{$t=1$ to $T+1$}
            \State $\boldsymbol{a}_t = \boldsymbol{A} \cdot \boldsymbol{z}_{t-1}$
            \State $\boldsymbol{z}_t = (\uparrow \boldsymbol{x}_t) \cap m(\boldsymbol{a}_t)$
            \State $\boldsymbol{A} = \boldsymbol{A} + \Delta \boldsymbol{A}$ \CComment{Update $\boldsymbol{A}$ via Eqn.~\ref{eqn:heb_A}}
            \State $\hat{\boldsymbol{z}}_t = \delta(\boldsymbol{A} \cdot \boldsymbol{z}_{t-1})$
            \State $\boldsymbol{B} = \boldsymbol{B} + \Delta \boldsymbol{B}$ \CComment{Update $\boldsymbol{B}$ via Eqn.~\ref{eqn:heb_B}}
        \EndFor
    \EndProcedure
    \vspace{38pt}
    \end{algorithmic}
\end{algorithm}
\end{minipage}
\hfill
\begin{minipage}[t]{0.485\textwidth}
\begin{algorithm}[H]
  \small
    \caption{
        : \textbf{Sequence Generation}. Given a noisy sequence (i.e., online), or the first input in a sequence (i.e., offline). The model uses the learned functions $\boldsymbol{A}$ and $\boldsymbol{B}$ to generate the full sequence. $\sim$ denotes sampled from (eqn~\ref{eqn:sample}).
    }
    \label{alg:recall}
    \begin{algorithmic}[1]
    \Procedure{Generate}{$\boldsymbol{x}_0$ or $\boldsymbol{x}$ }
    \State $\boldsymbol{z}_0 = (\uparrow \boldsymbol{x}_0) \cap m(\boldsymbol{a}_0)$ \CComment{Eqn.~\ref{eqn:context_embedding}}
    \For{$t=1$ to $T+1$}
        \State $\boldsymbol{a}_t = \boldsymbol{A} \cdot \boldsymbol{z}_{t-1}$
        \State $\hat{\boldsymbol{z}}_t = \delta(\boldsymbol{a}_t)$
        \State $\tilde{\boldsymbol{x}}_t = \begin{cases} \boldsymbol{x}_t \hspace{33pt} \text{online} \\ \sim (\downarrow \hat{\boldsymbol{z}}_t) \quad \text{offline} $\hspace{42pt} \CComment{Eqn.~\ref{eqn:sample}}$  \end{cases}$ 
        
        \For{i=1 to iters}\CComment{Attractors iterations}
            \State $\tilde{\boldsymbol{x}}_t = \delta(\boldsymbol{B} \cdot \tilde{\boldsymbol{x}}_t) \cap (\downarrow \hat{\boldsymbol{z}}_t)$
        \EndFor
        
        \State $\boldsymbol{z}_t = (\uparrow \tilde{\boldsymbol{x}}_t) \cap m(\boldsymbol{a}_t)$
    \EndFor

    \EndProcedure
    \end{algorithmic}
\end{algorithm}
\end{minipage}
\vspace{-1em}
\end{figure}

\textbf{State Transition Learning}\; The transition between latent states is learned through local computations with Hebbian-based rules. We modify the synaptic weights $\boldsymbol{A}$ to model the transition between pre-synaptic neurons $\boldsymbol{z}_{t-1}$ and post-synaptic neurons $\boldsymbol{z}_t$. Only the synapses with active pre-synaptic neurons are modified. The weights operate on context-dependant latent states (i.e., $\boldsymbol{z}_{t-1} \to \boldsymbol{z}_t$); thus, the learning of one transition does not overwrite previously learned transition of different contexts, regardless of the input contents (i.e., $\boldsymbol{x}_{t-1}$). We use the learning constant coefficients $\eta_A^+$ and $\eta_A^-$ to independently control learning and forgetting behavior, as shown in equation~\ref{eqn:heb_A}. A lower $\eta_A^-$ encourages learning multiple possibilities by slowing down the forgetting behavior. 
\begin{equation}\label{eqn:heb_A}
    \Delta \boldsymbol{A} = 
    \underbrace{\eta_A^+ \cdot \boldsymbol{z}_{t-1} \cdot {\boldsymbol{z}_{t}}^T}_{\Delta \boldsymbol{A}_{\text{increase}}}
    + 
    \underbrace{\eta_A^- \cdot \boldsymbol{z}_{t-1} \cdot {(1-\boldsymbol{z}_{t})}^T}_{\Delta \boldsymbol{A}_{\text{decrease}}}
\end{equation}

\textbf{Contrastive Attractors Formation}\; The attractors are formed in an online manner by contrasting the input observation $\boldsymbol{x}_t$ to the predicted union set of possibilities $\downarrow \hat{\boldsymbol{z}_t}$. The goal is to learn excitatory synapses between active neurons of $\boldsymbol{x}_t$, and bidirectional inhibitory synapses between $\boldsymbol{x}_t$ and the union set of predicted possibilities \textit{excluding} the $\boldsymbol{x}_t$ possibility (i.e., ${(\downarrow\hat{\boldsymbol{z}}_{t})} \setminus \boldsymbol{x}_{t})$), as shown in equation~\ref{eqn:heb_B}.
\begin{equation}\label{eqn:heb_B}
    \Delta \boldsymbol{B} = 
    \underbrace{\eta_B^+ \cdot \boldsymbol{x}_{t} \cdot {\boldsymbol{x}_{t}}^T}_{\Delta \boldsymbol{B}_{\text{increase}}}
    + 
    \underbrace{\eta_B^- \cdot [\boldsymbol{x}_{t} \cdot {((\downarrow\hat{\boldsymbol{z}}_{t})} \setminus \boldsymbol{x}_{t})^T  + {((\downarrow\hat{\boldsymbol{z}}_{t})} \setminus \boldsymbol{x}_{t}) \cdot \boldsymbol{x}_{t}^T ] }_{\Delta \boldsymbol{B}_{\text{decrease}}}
\end{equation}

\subsection{Sequence Generation}

\begin{figure}[h]
\centering
\includegraphics[width = \linewidth]{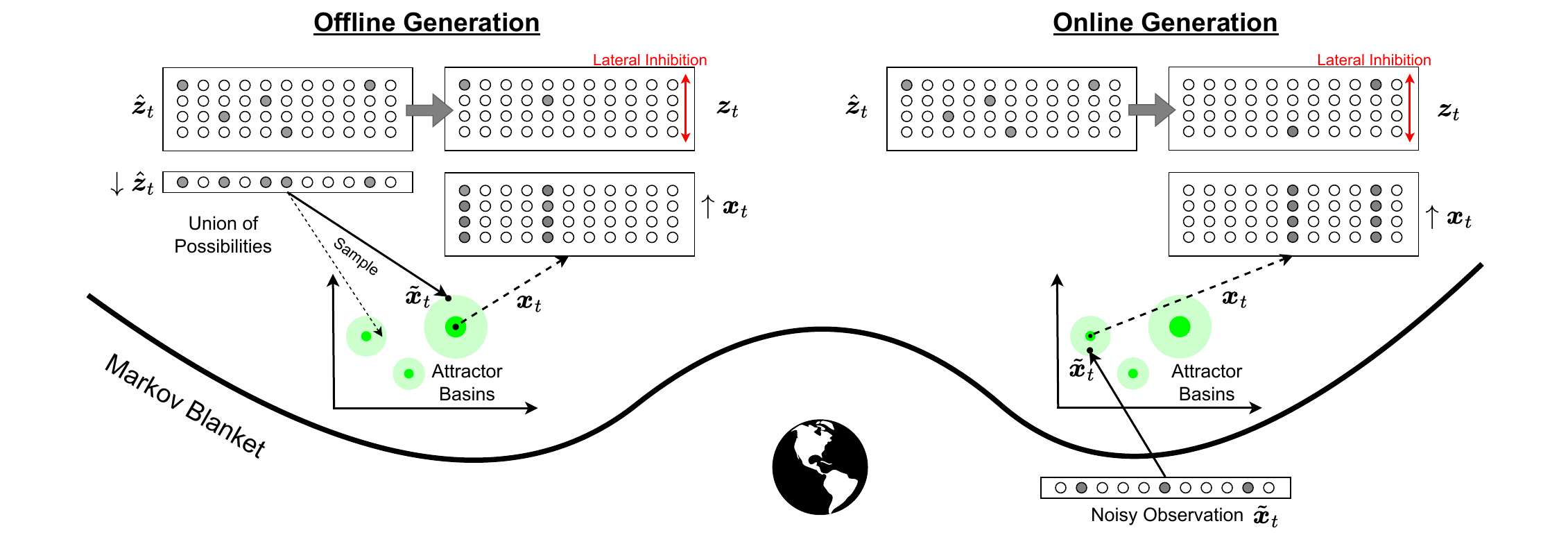}
\caption{Sequence Generation. (Left): Offline generation by sampling a single possibility (i.e., attractor point) from a union of predicted possibilities. (Right): Online generation by removing noise from an observation using the prior beliefs about the observed state. Markov Blanket separates the agent's latent variables from the world observable states.} \label{fig:generation}
\end{figure}

After learning one or multiple sequences using algorithm~\ref{alg:learn}, we use algorithm~\ref{alg:recall} to generate sequences. First, we define two generative tasks: online and offline. In online sequence generation, a noisy version of the sequence is provided as input, and the model is expected to generate the original learned sequence. In offline sequence generation, the model is only provided with the first input, and it is expected to generate the entire sequence auto-regressively. For cases with equally valid future predictions (e.g., ``a'' and ``e'' after ``TH'' in ``THAT'' and ``THEY''), the model is expected to stochastically generate either one of the possibilities (i.e., ``THAT'' or ``THEY''). The online generation task is a more challenging extension of the online recall task in~\cite{tang2024sequential}, where the noise-free inputs are provided, and the model only makes a 1-step prediction into the future. During offline sequence generation, the model randomly samples from the union set of predictions $\downarrow \hat{\boldsymbol{z}}$ a single SDR with $W$ active neurons (equation~\ref{eqn:sample}) to initialize the iterative attractor procedure. $\pi$ denotes a random permutation function. This random permutation function allows the model to randomly generate a different sequence with every generation.
\begin{equation}\label{eqn:sample}
    \tilde{\boldsymbol{x}}^i = 
    \begin{cases} 
    1 & \text{if } i \in \{\pi(I(\downarrow\hat{\boldsymbol{z}}_t))^w\}_{w=0}^{W}, \\ 
    0 & \text{otherwise} 
    \end{cases}
\end{equation}

  \section{Experiments}
\label{sec:experiments}

\subsection{Evaluation and Metrics}

\textbf{Metrics}\; To evaluate the similarity of two SDRs, we use the Jaccard Index (i.e., IoU), which focuses on the active bits in sparse binary representations. Since the sparsity $S$ of the representations can change across experiments and methods, we normalize the IoU by the expected IoU (Derived in Theorem~\ref{theorem:iou} in Appendix~\ref{app:iou}) of two random SDRs at their specified sparsities. The normalized IoU is computed as $\frac{\text{IoU} - \mathbb{E}[\text{IoU}]}{1 - \mathbb{E}[\text{IoU}]}$. We use the Backward Transfer~\cite{lopez2017gradient} metric in evaluating catastrophic forgetting. Mean Squared Error (MSE) is used to compare images for vision datasets.

\textbf{Datasets}\; We perform evaluations on synthetic and real datasets. The synthetic datasets allow us to manually control variables (e.g., sequence length, correlation, noise, input size) to better understand the models' behavior across various settings. Additionally, we evaluate on real datasets of various types (e.g., protein sequences, text, vision) to benchmark PAM's performance relative to other models on more challenging and real sequences. For all vision experiments, we use an SDR autoencoder to learn a mapping between images and SDRs (Details on the autoencoder are provided in Appendix~\ref{app:datasets}). We run all experiments for 10 different seeds and report the mean and standard deviation in all the figures. More experimental details and results are provided in Appendices~\ref{app:datasets}~and~\ref{app:experiments}.

\begin{figure}[h]
\centering
\includegraphics[width = \linewidth]{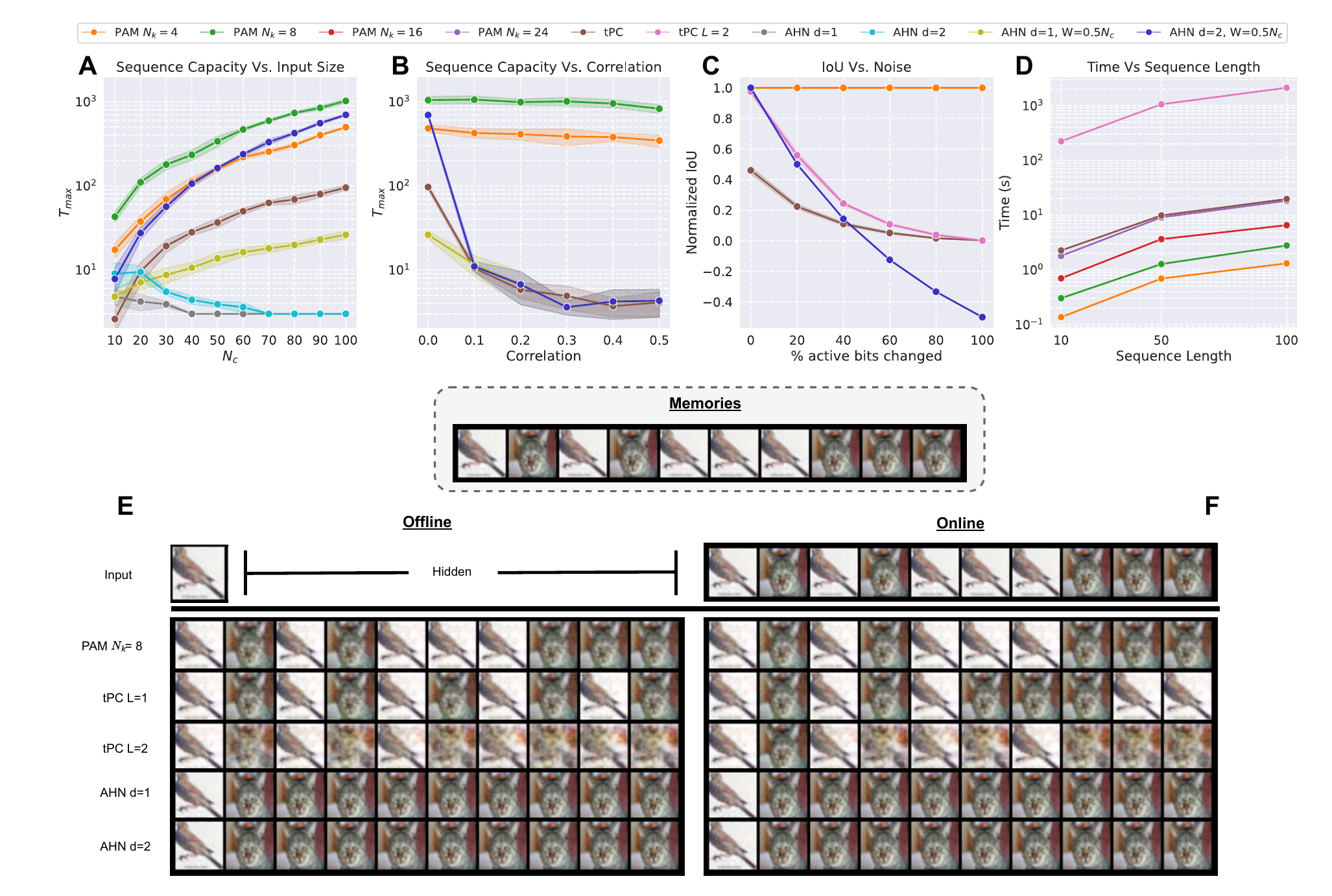}
\caption{Quantitative results on (\textbf{A-B}) Offline sequence capacity, (\textbf{C}) Noise robustness, and (\textbf{D}) Time of sequence learning and recall. Qualitative results on highly correlated CIFAR sequence in (\textbf{E}) offline and (\textbf{F}) online settings. The mean and standard deviation of 10 trials are reported for all plots.} \label{fig:results1}
\end{figure}

\subsection{Results}

We align our evaluation tasks with the desired characteristics of a biologically plausible sequence memory model, as listed in the introduction. We show that PAM outperforms current predictive coding and associative memory SoTA approaches on all tasks. Most importantly, PAM is capable of long context encoding, multiple possibilities generation, and learning continually and efficiently while avoiding catastrophic forgetting. These tasks pose numerous significant challenges to other methods.

\textbf{Offline Sequence Capacity}\; We evaluate the models' capacity to learn long sequences by varying the input size $N_c$, model parameters (e.g., $N_k$), and sequence correlation. The correlation is increased by reducing the number of unique patterns (i.e., vocab) used to create a sequence of length $T$. Correlation is computed as $1.0-\frac{\text{vocab}}{T}$. In Figure~\ref{fig:results1}~\textbf{A}, we vary the input size $N_c$ and ablate the models to find the maximum sequence length to be encoded and retrieved, in an \textit{offline} manner, with a Normalized IoU higher than 0.9. We set the number of active bits $W$ to be 5 unless otherwise specified. Results show that Hopfield Networks (HN) fail to learn with sparse representations; therefore, we use $W$ of $0.5N_c$ only for HN and normalize the IoU metric accordingly. PAM's capacity significantly increases with context neurons $N_k$, as expected. HN's capacity also increases with the polynomial degree $d$ of its separation function; however, as shown in Figure~\ref{fig:results1}~\textbf{B}, the capacity sharply drops as correlation increases. PAM retains its capacity with increasing correlation, reflecting its ability to encode context in long sequences (i.e., high-order Markov memory). This context encoding property is also demonstrated in the qualitative CIFAR~\cite{krizhevsky2009learning} results in Figure~\ref{fig:results1}~\textbf{E}~and~\textbf{F}, where a short sequence of images with high correlation is used. The model must uniquely encode the context to correctly predict at every step in the sequence. While PAM correctly predicts the full context, single layer tPC learns to indefinitely alternate between the patterns, while two-layered tPC attempts to average its predictions. AHN shows similar low performance and failure mode as in~\cite{tang2024sequential}.

\begin{figure}[h]
\centering
\includegraphics[width = \linewidth]{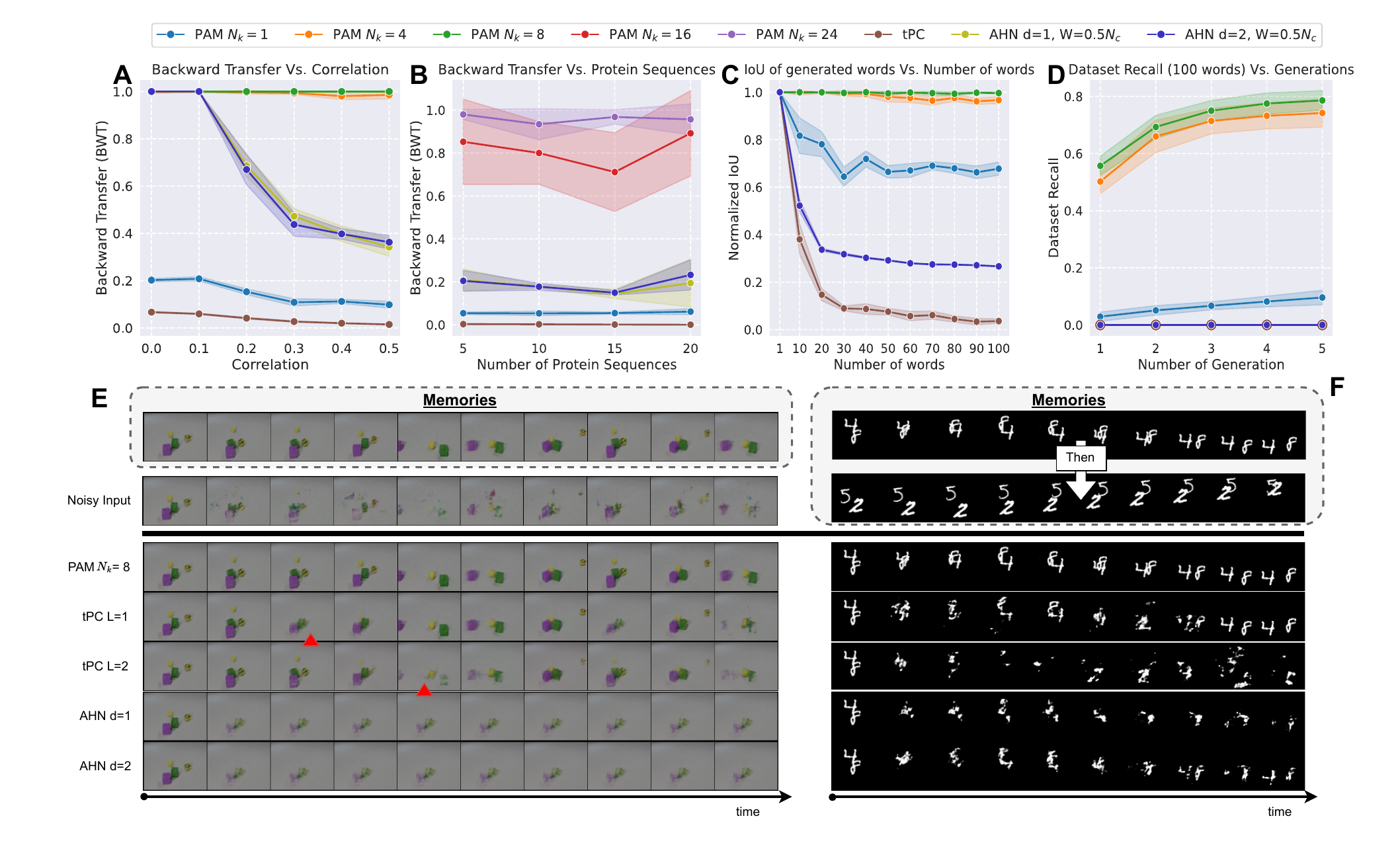}
\caption{Qualitative results on (\textbf{A}) synthetic and (\textbf{B}) protein sequences backward transfer, and (\textbf{C-D}) multiple possibilities generation on text datasets. Qualitative results on (\textbf{E}) noise robustness on CLEVRER sequence, and (\textbf{F}) catastrophic forgetting on Moving MNIST dataset. \protect\marksymbol{triangle*}{red} highlights the first frame with significant error. The mean and standard deviation of 10 trials are reported for all plots.}  \label{fig:results2}
\end{figure}

\textbf{Catastrophic Forgetting}\; To asses the model's performance in a continual learning setup, we sequentially train each model on multiple sequences and compute the Backward Transfer (BWT)~\cite{lopez2017gradient} metric by reporting the average normalized IoU on previously learned sequences after learning a new one. In Figure~\ref{fig:results2}~\textbf{A}, we report BWT for 50 synthetically generated sequences with varying correlation. AHN can avoid catastrophic forgetting when the patterns are not correlated, whereas tPC fails to retain previous knowledge regardless of the correlation value. PAM, with high enough context $N_k$, does not overwrite or forget previously learned sequences after learning new ones but performs poorly when $N_k=1$, as expected. We repeat the experiment on more challenging sequences from ProteinNet~\cite{alquraishi2019proteinnet}, which contains highly correlated ($>0.9$), and long, sequences (details in appendix). The results in Figure~\ref{fig:results2}~\textbf{B} show a similar trend with PAM requiring more context neurons $N_k$ to handle the more challenging data. Qualitative results on moving-MNIST~\cite{srivastava2015unsupervised} in Figure~\ref{fig:results2}~\textbf{F} further demonstrate the catastrophic forgetting challenge where the learning of the second sequence overwrites the learned sequence. PAM successfully retrieves the previously learned sequence while the other models fail.

\textbf{Multiple Possibilities Generation}\; In addition to accurately encoding input contexts, PAM is designed to represent multiple valid possibilities and sample a single possibility. We perform evaluation on a dataset of four-letter English words (details in appendix), which includes many possible future completions (e.g., ``th[is]'', ``th[at]'', ``th[em]'', etc.) We train PAM on the list of letters sequentially (i.e., one word at a time); the other methods are trained in a simpler batched setup as in~\cite{tang2024sequential} because they suffer from catastrophic forgetting. This puts PAM at a disadvantage, but as shown in Figure~\ref{fig:results2}~\textbf{C}, PAM still significantly outperforms the other methods in accurately generating valid words (high average IoU) in an offline manner. Both tPC and AHN fail to generate meaningful words when trained on sequences with multiple future possibilities. Figure~\ref{fig:results2}~\textbf{D} further demonstrates the stochastic generative property of PAM. We show PAM's ability to recall more of the dataset as it repeats the generation process, whereas PC and AHN fail in the dataset recall task.

\textbf{Online Noise Robustness}\; The online generation ability of PAM shown in Figure~\ref{fig:generation} allows the model to use the learned attractors to clean the noisy observations before using them as inputs to the predictor. This step allows the model to use its prior belief about future observations to modify the noisy inputs. In Figure~\ref{fig:results1}~\textbf{C}, we evaluate the models' performances by changing a percentage of the active bits during online generation. PAM is able to completely replace the noisy input with its prior belief if it does not exist in the predicted set of possibilities $\hat{\boldsymbol{z}}$. In contrast, the other methods use the noisy inputs, thus hindering their performances. We provide qualitative results on CLEVRER~\cite{yi2019clevrer} in Figure~\ref{fig:results2}~\textbf{E}; PAM retrieves the original sequence despite getting noisy inputs (40\% noise), and outperforms the other models. Interestingly, tPC performs reasonably well on this task despite the added noise.

\textbf{Efficiency}\; We report, in Figure~\ref{fig:results1}~\textbf{D}, the time each model requires to learn and recall a sequence. For this study, we use input size $N_c=100$ and vary the sequence length. PAM operates entirely on \textit{CPU}. The results show that a single-layer tPC model requires more time than all PAM variants ($N_k \leq 24$). Additionally, a two-layered tPC requires two to three orders of magnitude more time than PAM or single-layered tPC, significantly limiting its scalability and practicality when applied to real data with long sequences.

  \section{Conclusion}
\label{sec:conclusion}

We proposed \textit{PAM}, a biologically plausible generative model inspired by neuroscience findings and theories of cognition. We demonstrate that PAM is capable of encoding unique contexts with tremendous scalable capacity that is not affected by sequence correlation or noise. PAM is a generative model; it can represent multiple possibilities as a union of SDRs (a property of sparsity) and sample single possibilities, thus predicting multiple steps in the future despite multiple possible continuations. We also show that PAM does not suffer from catastrophic forgetting as it learns multiple sequences. PAM is trained using local computations through Hebbian rules and runs efficiently on CPUs. Future directions include hierarchical sensory processing and higher-order sparse predictive models.

  \clearpage
  {
    \small
    
    \section*{Acknowledgements} This research was supported by the US National Science Foundation Grants CNS 1513126 and IIS 1956050.
    
    \bibliographystyle{plain}
    \bibliography{references.bib}

\begin{thebibliography}{10}

\bibitem{aakur2020action}
Sathyanarayanan Aakur and Sudeep Sarkar.
\newblock Action localization through continual predictive learning.
\newblock In {\em Computer Vision--ECCV 2020: 16th European Conference,
  Glasgow, UK, August 23--28, 2020, Proceedings, Part XIV 16}, pages 300--317.
  Springer, 2020.

\bibitem{aakur2019perceptual}
Sathyanarayanan~N Aakur and Sudeep Sarkar.
\newblock A perceptual prediction framework for self supervised event
  segmentation.
\newblock In {\em Proceedings of the IEEE/CVF Conference on Computer Vision and
  Pattern Recognition}, pages 1197--1206, 2019.

\bibitem{afraimovich2015sequential}
Valentin Afraimovich, Xue Gong, and Mikhail Rabinovich.
\newblock Sequential memory: Binding dynamics.
\newblock {\em Chaos: An Interdisciplinary Journal of Nonlinear Science},
  25(10), 2015.

\bibitem{ahmad2015properties}
Subutai Ahmad and Jeff Hawkins.
\newblock Properties of sparse distributed representations and their
  application to hierarchical temporal memory.
\newblock {\em arXiv preprint arXiv:1503.07469}, 2015.

\bibitem{ahmad2016neurons}
Subutai Ahmad and Jeff Hawkins.
\newblock How do neurons operate on sparse distributed representations? a
  mathematical theory of sparsity, neurons and active dendrites.
\newblock {\em arXiv preprint arXiv:1601.00720}, 10, 2016.

\bibitem{ahmad2017unsupervised}
Subutai Ahmad, Alexander Lavin, Scott Purdy, and Zuha Agha.
\newblock Unsupervised real-time anomaly detection for streaming data.
\newblock {\em Neurocomputing}, 262:134--147, 2017.

\bibitem{alquraishi2019proteinnet}
Mohammed AlQuraishi.
\newblock Proteinnet: a standardized data set for machine learning of protein
  structure.
\newblock {\em BMC bioinformatics}, 20:1--10, 2019.

\bibitem{assran2023self}
Mahmoud Assran, Quentin Duval, Ishan Misra, Piotr Bojanowski, Pascal Vincent,
  Michael Rabbat, Yann LeCun, and Nicolas Ballas.
\newblock Self-supervised learning from images with a joint-embedding
  predictive architecture.
\newblock In {\em Proceedings of the IEEE/CVF Conference on Computer Vision and
  Pattern Recognition}, pages 15619--15629, 2023.

\bibitem{bardes2024revisiting}
Adrien Bardes, Quentin Garrido, Jean Ponce, Xinlei Chen, Michael Rabbat, Yann
  LeCun, Mahmoud Assran, and Nicolas Ballas.
\newblock Revisiting feature prediction for learning visual representations
  from video.
\newblock {\em arXiv preprint arXiv:2404.08471}, 2024.

\bibitem{bardes2023mc}
Adrien Bardes, Jean Ponce, and Yann LeCun.
\newblock Mc-jepa: A joint-embedding predictive architecture for
  self-supervised learning of motion and content features.
\newblock {\em arXiv preprint arXiv:2307.12698}, 2023.

\bibitem{caron2021emerging}
Mathilde Caron, Hugo Touvron, Ishan Misra, Herv{\'e} J{\'e}gou, Julien Mairal,
  Piotr Bojanowski, and Armand Joulin.
\newblock Emerging properties in self-supervised vision transformers.
\newblock In {\em Proceedings of the IEEE/CVF international conference on
  computer vision}, pages 9650--9660, 2021.

\bibitem{chaudhry2024long}
Hamza Chaudhry, Jacob Zavatone-Veth, Dmitry Krotov, and Cengiz Pehlevan.
\newblock Long sequence hopfield memory.
\newblock {\em Advances in Neural Information Processing Systems}, 36, 2024.

\bibitem{chen2021exploring}
Xinlei Chen and Kaiming He.
\newblock Exploring simple siamese representation learning.
\newblock In {\em Proceedings of the IEEE/CVF conference on computer vision and
  pattern recognition}, pages 15750--15758, 2021.

\bibitem{conway2009episodic}
Martin~A Conway.
\newblock Episodic memories.
\newblock {\em Neuropsychologia}, 47(11):2305--2313, 2009.

\bibitem{elias1955predictive}
Peter Elias.
\newblock Predictive coding--i.
\newblock {\em IRE transactions on information theory}, 1(1):16--24, 1955.

\bibitem{friston2010free}
Karl Friston.
\newblock The free-energy principle: a unified brain theory?
\newblock {\em Nature reviews neuroscience}, 11(2):127--138, 2010.

\bibitem{friston2007variational}
Karl Friston, J{\'e}r{\'e}mie Mattout, Nelson Trujillo-Barreto, John Ashburner,
  and Will Penny.
\newblock Variational free energy and the laplace approximation.
\newblock {\em Neuroimage}, 34(1):220--234, 2007.

\bibitem{friston2008variational}
Karl~J Friston, N~Trujillo-Barreto, and Jean Daunizeau.
\newblock Dem: a variational treatment of dynamic systems.
\newblock {\em Neuroimage}, 41(3):849--885, 2008.

\bibitem{grill2020bootstrap}
Jean-Bastien Grill, Florian Strub, Florent Altch{\'e}, Corentin Tallec, Pierre
  Richemond, Elena Buchatskaya, Carl Doersch, Bernardo Avila~Pires, Zhaohan
  Guo, Mohammad Gheshlaghi~Azar, et~al.
\newblock Bootstrap your own latent-a new approach to self-supervised learning.
\newblock {\em Advances in neural information processing systems},
  33:21271--21284, 2020.

\bibitem{haggard1998planning}
Patrick Haggard.
\newblock Planning of action sequences.
\newblock {\em Acta Psychologica}, 99(2):201--215, 1998.

\bibitem{han2018deep}
Kuan Han, Haiguang Wen, Yizhen Zhang, Di~Fu, Eugenio Culurciello, and Zhongming
  Liu.
\newblock Deep predictive coding network with local recurrent processing for
  object recognition.
\newblock {\em Advances in neural information processing systems}, 31, 2018.

\bibitem{han2019video}
Tengda Han, Weidi Xie, and Andrew Zisserman.
\newblock Video representation learning by dense predictive coding.
\newblock In {\em Proceedings of the IEEE/CVF International Conference on
  Computer Vision Workshops}, pages 0--0, 2019.

\bibitem{hawkins2016neurons}
Jeff Hawkins and Subutai Ahmad.
\newblock Why neurons have thousands of synapses, a theory of sequence memory
  in neocortex.
\newblock {\em Frontiers in neural circuits}, 10:174222, 2016.

\bibitem{hopfield1982neural}
John~J Hopfield.
\newblock Neural networks and physical systems with emergent collective
  computational abilities.
\newblock {\em Proceedings of the national academy of sciences},
  79(8):2554--2558, 1982.

\bibitem{hosoya2005dynamic}
Toshihiko Hosoya, Stephen~A Baccus, and Markus Meister.
\newblock Dynamic predictive coding by the retina.
\newblock {\em Nature}, 436(7047):71--77, 2005.

\bibitem{kage2023implementing}
Hiroshi Kage.
\newblock Implementing associative memories by echo state network for the
  applications of natural language processing.
\newblock {\em Machine Learning with Applications}, 11:100449, 2023.

\bibitem{kanerva1988sparse}
Pentti Kanerva.
\newblock {\em Sparse distributed memory}.
\newblock MIT press, 1988.

\bibitem{koller2009probabilistic}
Daphne Koller and Nir Friedman.
\newblock {\em Probabilistic graphical models: principles and techniques}.
\newblock MIT press, 2009.

\bibitem{kosko1988bidirectional}
Bart Kosko.
\newblock Bidirectional associative memories.
\newblock {\em IEEE Transactions on Systems, man, and Cybernetics},
  18(1):49--60, 1988.

\bibitem{krizhevsky2009learning}
Alex Krizhevsky, Geoffrey Hinton, et~al.
\newblock Learning multiple layers of features from tiny images.
\newblock 2009.

\bibitem{krotov2016dense}
Dmitry Krotov and John~J Hopfield.
\newblock Dense associative memory for pattern recognition.
\newblock {\em Advances in neural information processing systems}, 29, 2016.

\bibitem{lauritzen1996graphical}
Steffen~L Lauritzen.
\newblock {\em Graphical models}, volume~17.
\newblock Clarendon Press, 1996.

\bibitem{lecun2022path}
Yann LeCun.
\newblock A path towards autonomous machine intelligence version 0.9. 2,
  2022-06-27.
\newblock 2022.

\bibitem{lopez2017gradient}
David Lopez-Paz and Marc'Aurelio Ranzato.
\newblock Gradient episodic memory for continual learning.
\newblock {\em Advances in neural information processing systems}, 30, 2017.

\bibitem{lotter2016deep}
William Lotter, Gabriel Kreiman, and David Cox.
\newblock Deep predictive coding networks for video prediction and unsupervised
  learning.
\newblock {\em arXiv preprint arXiv:1605.08104}, 2016.

\bibitem{millidge2022universal}
Beren Millidge, Tommaso Salvatori, Yuhang Song, Thomas Lukasiewicz, and Rafal
  Bogacz.
\newblock Universal hopfield networks: A general framework for single-shot
  associative memory models.
\newblock In {\em International Conference on Machine Learning}, pages
  15561--15583. PMLR, 2022.

\bibitem{mounir2022self}
Ramy Mounir, Sathyanarayanan Aakur, and Sudeep Sarkar.
\newblock Self-supervised temporal event segmentation inspired by cognitive
  theories.
\newblock In {\em Advanced Methods and Deep Learning in Computer Vision}, pages
  405--448. Elsevier, 2022.

\bibitem{mounir2021spatio}
Ramy Mounir, Roman Gula, J{\"o}rn Theuerkauf, and Sudeep Sarkar.
\newblock Spatio-temporal event segmentation for wildlife extended videos.
\newblock In {\em International Conference on Computer Vision and Image
  Processing}, pages 48--59. Springer, 2021.

\bibitem{mounir2023towards}
Ramy Mounir, Ahmed Shahabaz, Roman Gula, J{\"o}rn Theuerkauf, and Sudeep
  Sarkar.
\newblock Towards automated ethogramming: Cognitively-inspired event
  segmentation for streaming wildlife video monitoring.
\newblock {\em International journal of computer vision}, 131(9):2267--2297,
  2023.

\bibitem{mounir2024streamer}
Ramy Mounir, Sujal Vijayaraghavan, and Sudeep Sarkar.
\newblock Streamer: Streaming representation learning and event segmentation in
  a hierarchical manner.
\newblock {\em Advances in Neural Information Processing Systems}, 36, 2024.

\bibitem{o1971entropy}
J~O'Neal.
\newblock Entropy coding in speech and television differential pcm systems
  (corresp.).
\newblock {\em IEEE Transactions on Information Theory}, 17(6):758--761, 1971.

\bibitem{rabiner1986introduction}
Lawrence Rabiner and Biinghwang Juang.
\newblock An introduction to hidden markov models.
\newblock {\em ieee assp magazine}, 3(1):4--16, 1986.

\bibitem{ramsauer2020hopfield}
Hubert Ramsauer, Bernhard Sch{\"a}fl, Johannes Lehner, Philipp Seidl, Michael
  Widrich, Lukas Gruber, Markus Holzleitner, Thomas Adler, David Kreil,
  Michael~K Kopp, et~al.
\newblock Hopfield networks is all you need.
\newblock In {\em International Conference on Learning Representations}, 2020.

\bibitem{rao1999predictive}
Rajesh~PN Rao and Dana~H Ballard.
\newblock Predictive coding in the visual cortex: a functional interpretation
  of some extra-classical receptive-field effects.
\newblock {\em Nature neuroscience}, 2(1):79--87, 1999.

\bibitem{rolls2010computational}
Edmund~T Rolls.
\newblock A computational theory of episodic memory formation in the
  hippocampus.
\newblock {\em Behavioural brain research}, 215(2):180--196, 2010.

\bibitem{salvatori2023brain}
Tommaso Salvatori, Ankur Mali, Christopher~L Buckley, Thomas Lukasiewicz,
  Rajesh~PN Rao, Karl Friston, and Alexander Ororbia.
\newblock Brain-inspired computational intelligence via predictive coding.
\newblock {\em arXiv preprint arXiv:2308.07870}, 2023.

\bibitem{salvatori2022learning}
Tommaso Salvatori, Luca Pinchetti, Beren Millidge, Yuhang Song, Tianyi Bao,
  Rafal Bogacz, and Thomas Lukasiewicz.
\newblock Learning on arbitrary graph topologies via predictive coding.
\newblock {\em Advances in neural information processing systems},
  35:38232--38244, 2022.

\bibitem{salvatori2021associative}
Tommaso Salvatori, Yuhang Song, Yujian Hong, Lei Sha, Simon Frieder, Zhenghua
  Xu, Rafal Bogacz, and Thomas Lukasiewicz.
\newblock Associative memories via predictive coding.
\newblock {\em Advances in Neural Information Processing Systems},
  34:3874--3886, 2021.

\bibitem{salvatori2022incremental}
Tommaso Salvatori, Yuhang Song, Beren Millidge, Zhenghua Xu, Lei Sha, Cornelius
  Emde, Rafal Bogacz, and Thomas Lukasiewicz.
\newblock Incremental predictive coding: A parallel and fully automatic
  learning algorithm.
\newblock {\em arXiv preprint arXiv:2212.00720}, 2022.

\bibitem{sompolinsky1986temporal}
Haim Sompolinsky and Ido Kanter.
\newblock Temporal association in asymmetric neural networks.
\newblock {\em Physical review letters}, 57(22):2861, 1986.

\bibitem{srinivasan1982predictive}
Mandyam~Veerambudi Srinivasan, Simon~Barry Laughlin, and Andreas Dubs.
\newblock Predictive coding: a fresh view of inhibition in the retina.
\newblock {\em Proceedings of the Royal Society of London. Series B. Biological
  Sciences}, 216(1205):427--459, 1982.

\bibitem{srivastava2015unsupervised}
Nitish Srivastava, Elman Mansimov, and Ruslan Salakhudinov.
\newblock Unsupervised learning of video representations using lstms.
\newblock In {\em International conference on machine learning}, pages
  843--852. PMLR, 2015.

\bibitem{tang2024sequential}
Mufeng Tang, Helen Barron, and Rafal Bogacz.
\newblock Sequential memory with temporal predictive coding.
\newblock {\em Advances in Neural Information Processing Systems}, 36, 2024.

\bibitem{tang2023recurrent}
Mufeng Tang, Tommaso Salvatori, Beren Millidge, Yuhang Song, Thomas
  Lukasiewicz, and Rafal Bogacz.
\newblock Recurrent predictive coding models for associative memory employing
  covariance learning.
\newblock {\em PLoS computational biology}, 19(4):e1010719, 2023.

\bibitem{tulving1972episodic}
Endel Tulving et~al.
\newblock Episodic and semantic memory.
\newblock {\em Organization of memory}, 1(381-403):1, 1972.

\bibitem{welch1995introduction}
Greg Welch, Gary Bishop, et~al.
\newblock An introduction to the kalman filter.
\newblock 1995.

\bibitem{whittington2017approximation}
James~CR Whittington and Rafal Bogacz.
\newblock An approximation of the error backpropagation algorithm in a
  predictive coding network with local hebbian synaptic plasticity.
\newblock {\em Neural computation}, 29(5):1229--1262, 2017.

\bibitem{yi2019clevrer}
Kexin Yi, Chuang Gan, Yunzhu Li, Pushmeet Kohli, Jiajun Wu, Antonio Torralba,
  and Joshua~B Tenenbaum.
\newblock Clevrer: Collision events for video representation and reasoning.
\newblock In {\em International Conference on Learning Representations}, 2019.

\bibitem{yoo2022bayespcn}
Jinsoo Yoo and Frank Wood.
\newblock Bayespcn: A continually learnable predictive coding associative
  memory.
\newblock {\em Advances in Neural Information Processing Systems},
  35:29903--29914, 2022.

\bibitem{zbontar2021barlow}
Jure Zbontar, Li~Jing, Ishan Misra, Yann LeCun, and St{\'e}phane Deny.
\newblock Barlow twins: Self-supervised learning via redundancy reduction.
\newblock In {\em International conference on machine learning}, pages
  12310--12320. PMLR, 2021.

\end{thebibliography}
  }

  \clearpage
  \newpage

\appendix
\begin{center}
    \hrule height 4pt
    \vspace{0.5\baselineskip}
    \huge\textbf{Supplementary Material}
    \vspace{0.5\baselineskip}
    \hrule height 1pt
\end{center}
\label{sec:appendix}

\section{Limitations}

PAM requires that its representations be sparse and binary (i.e., SDRs) in order to represent multiple possibilities as a union of SDRs with minimal overlap. Therefore, PAM cannot be directly applied to images in the input space like Dense Hopfield Networks. We argue that the neocortex encodes sensory input into SDRs for processing and does not operate directly on the input in its dense representation. Methods that operate directly on dense representations (e.g., images) are arguably less biologically plausible as the neocortex uses sparse binary representations (i.e., cell assemblies with a small fraction of firing neurons) to store and manipulate information. This paper focuses on learning multiple sequences without catastrophic forgetting and stochastically generating multiple possibilities efficiently, and we assume the sequence is provided in SDR format. Additionally, methods that operate on the input directly face challenges when the input is naturally sparse (see Figure~\ref{fig:results1}~\textbf{A}). Therefore, it is useful to encode the input into a representation with fixed sparsity before applying sequential memory learning algorithms. In future work, we plan to investigate how to encode high dimensional complex inputs (e.g., images) in a compositional part-whole structure of SDRs where we can apply PAM at different levels of abstraction.

\section{Theorems and Derivations}

\subsection{Variational Free Energy}\label{app:VFE}

\paragraph{Predictive Coding}

Consider a hierarchical generative model with hidden states $\{\boldsymbol{z}\}_{l=0}^{L}$, where $l \leq L$ denotes the level in the hierarchy. The conditional probability $p(\boldsymbol{z}_{l} | \boldsymbol{z}_{l+1})$ is assumed to be a multivariate Gaussian distribution with its mean calculated as a function $f_{l+1}$ of the higher-level hidden representation $\boldsymbol{z}_{l+1}$ and covariance $\boldsymbol{\Sigma}_l$ as shown in equation~\ref{eqn:PC-1}.

\begin{equation}\label{eqn:PC-1}
    p(\boldsymbol{z}_{l} | \boldsymbol{z}_{l+1}) = \mathcal{N}(f_{l+1}(\boldsymbol{z}_{l+1}), \boldsymbol{\Sigma}_l)
\end{equation}

The goal is to calculate the posterior of hidden states given an observation $\boldsymbol{x}$, (i.e., $P(\{\boldsymbol{z}\}_{l=0}^{L} | \boldsymbol{x})$). Since the prediction function $f$ contains a non-linear activation, we cannot analytically compute the posterior and we have to approximate it with a surrogate posterior (i.e., $q(\{\boldsymbol{z}\}_{l=0}^{L})$ by maximizing the Evidence Lower Bound (ELBO). We apply the mean field approximation to factorize this joint posterior probability into conditionally independent posteriors $\{q(\boldsymbol{z}_l)\}_{l=0}^{L}\}$, and apply the Laplace approximation to use Gaussian forms for the approximate distribution~\cite{friston2008variational, friston2007variational, salvatori2023brain}. Through these approximations, we can maximize the ELBO, or equivalently minimize the Variational Free Energy, in equation~\ref{eqn:PC-2}. 

\begin{equation}\label{eqn:PC-2}
    \underbrace{
    \mathbb{E}_{\boldsymbol{z} \sim q(\boldsymbol{z})} [\log ( \frac{q(\boldsymbol{z})} {p(\boldsymbol{x}, \boldsymbol{z})})]}_{\text{Variational Free Energy}}
    =
    \underbrace{
    \mathbb{E}_{\boldsymbol{z} \sim q(\boldsymbol{z})} [\log ( \frac{q(\boldsymbol{z})} {p(\boldsymbol{z})})]}_{D_{KL}(q||p)}
    +
    \underbrace{
    \mathbb{E}_{\boldsymbol{z} \sim q(\boldsymbol{z})} [\log ( \frac{1} {p(\boldsymbol{x}| \boldsymbol{z})})]}_{\text{Accuracy}}
\end{equation}

The variational free energy can be reduced to minimizing the negative log-likelihood (Accuracy term), which is simply the prediction error when the likelihood is assumed to take a Gaussian Form. Therefore, minimizing the prediction error reduces to the sum of the squared prediction error of every neuron.

\begin{derivation}\label{der:VFE-PC}
    Variational Free Energy derivation for the predictive coding objective function in equation~\ref{eqn:PC-2}. We approximate the true posterior $p(\boldsymbol{z}|\boldsymbol{x})$ with a surrogate posterior $q(\boldsymbol{z})$. The objective is to minimize the reverse KL divergence $D_{KL}(q(\boldsymbol{z})||p(\boldsymbol{z}|\boldsymbol{x}))$.

    \begin{align*}
    D_{KL}(q(\boldsymbol{z})||p(\boldsymbol{z}|\boldsymbol{x}))  &= \mathbb{E}_{\boldsymbol{z} \sim q(\boldsymbol{z})} [\log \frac{q(\boldsymbol{z})}{p(\boldsymbol{z}|\boldsymbol{x})}] && \text{(KL Divergence definition)}\\
    &= \mathbb{E}_{\boldsymbol{z} \sim q(\boldsymbol{z})} [\log \frac{q(\boldsymbol{z}) p(\boldsymbol{x})}{ p(\boldsymbol{x}|\boldsymbol{z}) p(\boldsymbol{z})}] && \text{(Bayes Theorem)}\\
    &= \mathbb{E}_{\boldsymbol{z} \sim q(\boldsymbol{z})} [\log \frac{q(\boldsymbol{z})}{ p(\boldsymbol{x}|\boldsymbol{z}) p(\boldsymbol{z})}] +\mathbb{E}_{\boldsymbol{z} \sim q(\boldsymbol{z})} [\log p(\boldsymbol{x}) ]  && \text{(Linearity of expectations)}\\
    D_{KL}(q(\boldsymbol{z})||p(\boldsymbol{z}|\boldsymbol{x})) &= \underbrace{\mathbb{E}_{\boldsymbol{z} \sim q(\boldsymbol{z})} [\log \frac{q(\boldsymbol{z})}{ p(\boldsymbol{x}|\boldsymbol{z}) p(\boldsymbol{z})}]}_{\text{Variational Free Energy}} + \log p(\boldsymbol{x})  && \text{(Evidence does not depend on $q(\boldsymbol{x})$)}
    \end{align*}

    To minimize the KL divergence, we can minimize the Variational Free energy instead because the Evidence term ($\log p(\boldsymbol{x})$) is constant negative term. The Variational Free Energy can be further simplified as follows:

    \begin{align*}
    \mathbb{E}_{\boldsymbol{z} \sim q(\boldsymbol{z})} [\log \frac{q(\boldsymbol{z})}{ p(\boldsymbol{x}|\boldsymbol{z}) p(\boldsymbol{z})}] &= 
    \mathbb{E}_{\boldsymbol{z} \sim q(\boldsymbol{z})} [\log \frac{1}{ p(\boldsymbol{x}|\boldsymbol{z})}]
    +
    \mathbb{E}_{\boldsymbol{z} \sim q(\boldsymbol{z})} [\log \frac{q(\boldsymbol{z})}{p(\boldsymbol{z})}]
    && \text{(Linearity of Expectations)}\\
    \underbrace{\mathbb{E}_{\boldsymbol{z} \sim q(\boldsymbol{z})} [\log \frac{q(\boldsymbol{z})}{ p(\boldsymbol{x}|\boldsymbol{z}) p(\boldsymbol{z})}]}_{\text{Variational Free Energy}} 
    &= 
    \underbrace{\mathbb{E}_{\boldsymbol{z} \sim q(\boldsymbol{z})} [\log \frac{1}{ p(\boldsymbol{x}|\boldsymbol{z})}]}_{\text{Error}}
    +
    D_{KL}(q(\boldsymbol{z})||p(\boldsymbol{z}))
    && \text{(KL Divergence definition)}
    \end{align*}

    We arrive at equation~\ref{eqn:PC-2}. Minimizing the error term (i.e., negative log-likelihood) is equivalent to minimizing the Sum of Squared Error (SSE) when a Gaussian form is assumed for the likelihood $p(\boldsymbol{x}|\boldsymbol{z})$.
    
\end{derivation}

\begin{derivation}\label{der:VFE-PAM}
    Variational Free Energy derivation for a State Space Model (SSM) in equation~\ref{eqn:pam-vfe}. Latent states are denoted with $\boldsymbol{z}$, whereas observations are denoted with $\boldsymbol{x}$. We assume non-linear transition and emission function (i.e., $f$ and $g$), therefore a variational approximation is needed to approximate the true posterior $p(\boldsymbol{z}_t|\boldsymbol{z_{t-1}}, \boldsymbol{x}_t)$ with a surrogate posterior $q(\boldsymbol{z}_t)$. As in derivation~\ref{der:VFE-PC}, the goal is to minimize the divergence between the true posterior and the approximate posterior (i.e., $D_{KL}(q(\boldsymbol{z}_t)||p(\boldsymbol{z}_t|\boldsymbol{z_{t-1}}, \boldsymbol{x}_t))$). Note that, for notation brevity, $\mathbb{E}_q \equiv \mathbb{E}_{\boldsymbol{z} \sim q(\boldsymbol{z}_t)}$.

    \begin{align*}
    D_{KL}(q(\boldsymbol{z}_t)||p(\boldsymbol{z}_t|\boldsymbol{z_{t-1}}, \boldsymbol{x}_t))  &= \mathbb{E}_q [\log \frac{q(\boldsymbol{z}_t)}{p(\boldsymbol{z}_t|\boldsymbol{z_{t-1}}, \boldsymbol{x}_t)}] && \text{(KL Divergence definition)}\\
    &= \mathbb{E}_q [\log \frac{q(\boldsymbol{z}_t) p(\boldsymbol{x}_t | \boldsymbol{z}_{t-1})}{ p(\boldsymbol{x}_t|\boldsymbol{z}_t, \boldsymbol{z}_{t-1}) p(\boldsymbol{z}_t|\boldsymbol{z}_{t-1})}] && \text{(Bayes Theorem)}\\
    &= \mathbb{E}_q [\log \frac{q(\boldsymbol{z}_t) p(\boldsymbol{x}_t | \boldsymbol{z}_{t-1})}{ p(\boldsymbol{x}_t|\boldsymbol{z}_t, \cancel{\boldsymbol{z}_{t-1}}) p(\boldsymbol{z}_t|\boldsymbol{z}_{t-1})}] && \text{(Conditional Independence)}\\
    D_{KL}(q(\boldsymbol{z}_t)||p(\boldsymbol{z}_t|\boldsymbol{z_{t-1}}, \boldsymbol{x}_t)) &= \underbrace{\mathbb{E}_q [\log \frac{q(\boldsymbol{z}_t)}{ p(\boldsymbol{x}_t|\boldsymbol{z}_t) p(\boldsymbol{z}_t|\boldsymbol{z}_{t-1})}]}_{\text{Variational Free Energy}} + \mathbb{E}_q [\log p(\boldsymbol{x}_t|\boldsymbol{z}_{t-1}) ]  && \text{(Linearity of expectations)}
    \end{align*}

    We can minimize the Variational Free Energy term which reduces to log-likelihood of two prediction error terms and the negative entropy of the approximate posterior $q(\boldsymbol{z}_t)$ as shown below.

    \begin{equation*}
         \mathbb{E}_q [\log \frac{q(\boldsymbol{z}_t)}{ p(\boldsymbol{x}_t|\boldsymbol{z}_t) p(\boldsymbol{z}_t|\boldsymbol{z}_{t-1})}]
        = 
        \underbrace{\mathbb{E}_q [\log \frac{1}{p(\boldsymbol{z}_t|\boldsymbol{z}_{t-1})}]}_{\text{Latent State Error}}
        +
        \underbrace{\mathbb{E}_q [\log \frac{1}{ p(\boldsymbol{x}_t|\boldsymbol{z}_t)}]}_{\text{Observation Error}}
        -
        \underbrace{\mathbb{E}_q [\log \frac{1}{q(\boldsymbol{z}_t)}]}_{\text{entropy } \mathcal{H}_q}
    \end{equation*}

    Minimizing the Variational Free Energy above forces $q(\boldsymbol{z}_t)$ to better approximate the true posterior.

\end{derivation}

\subsection{Gaussian Mixture Model and Hopfield Recall}~\label{app:GMM}

\begin{derivation}\label{der:GMM}
    \newcommand{\bx}{\boldsymbol{x}}
    \newcommand{\bz}{\boldsymbol{z}}
    \newcommand{\ddx}{\frac{\partial}{\partial \bx}}
    \newcommand{\gm}{\mathcal{N}(\bx; \boldsymbol{\mu}_c, \boldsymbol{\Sigma}_c) }
    \newcommand{\cs}{\sum_{c=1}^{C} w_c }
    \newcommand{\gmm}{ \cs \cdot \gm}
    \newcommand{\dgmm}{ - \boldsymbol{\Sigma}_c^{-1} (\bx-\boldsymbol{\mu}_c) }

    We derive theorem~\ref{theorem:gmm} which states that the maximization of the log-likelihood of $p(\boldsymbol{x}_t|\boldsymbol{z}_t)$ in the form of a Gaussian Mixture Model is equivalent to the recall function in Hopfield networks (i.e., eqn~\ref{eqn:UHN-cases}), where the means of the GMM (i.e., $\{\boldsymbol{\mu}_c\}_{c=1}^C$) represents the attractors of a Hopfield model. To maximize the log-likelihood, we compute its partial derivative with respect to $\boldsymbol{x}$.

    \begin{align*}
        \ddx \log p(\bx|\bz) &= \ddx [\log \gmm]\\
        &= \frac{1}{\gmm} \cdot \ddx \gmm\\
        &= \frac{1}{\gmm} \cdot \cs \ddx \gm\\
        &= \frac{1}{\gmm} \cdot \cs \cdot [\dgmm] \cdot \gm\\
        &= \frac{\gmm \cdot \dgmm}{\gmm}\\
    \end{align*}

    By setting the partial derivative of the log-likelihood to $0$, we can estimate the value of $\boldsymbol{x}$ which maximizes the function $\log p(\bx|\bz)$.

    \begin{align*}
        \ddx \log p(\bx|\bz) &= 0\\
        \frac{\gmm \cdot \dgmm}{\gmm} &= 0\\
        \frac{\gmm \cdot \boldsymbol{\Sigma}_c^{-1}\bx}{\cancel{\gmm}} &=  \frac{\gmm \cdot \boldsymbol{\Sigma}_c^{-1}\boldsymbol{\mu}_c}{\cancel{\gmm}}\\
    \end{align*}

    Finally, we can rearrange the equation in terms of $\bx$ and show that it is equivalent to the Hopfield recall function where the recall value $\bx$ equals a weighted average of the attractors (i.e., means of GMM $\{\boldsymbol{\mu}_c\}_{c=1}^C$), with the weights being a similarity score function.

    \begin{align*}
        \bx &= \frac{\gmm \cdot \boldsymbol{\Sigma}_c^{-1}\boldsymbol{\mu}_c}{\gmm \cdot \boldsymbol{\Sigma}_c^{-1}}\\
        \bx &= \sum_{c=1}^C  \underbrace{\frac{w_c \cdot \gm \cdot \boldsymbol{\Sigma}_c^{-1}}{\gmm \cdot \boldsymbol{\Sigma}_c^{-1}}}_{\text{similarity score}}
        \cdot
        \underbrace{\boldsymbol{\mu}_c}_{\text{projection}}\\
    \end{align*}

\end{derivation}

\subsection{Expected IoU of Random SDRs}\label{app:iou}

\begin{figure}[h]
\centering
\includegraphics[width = 0.7\linewidth]{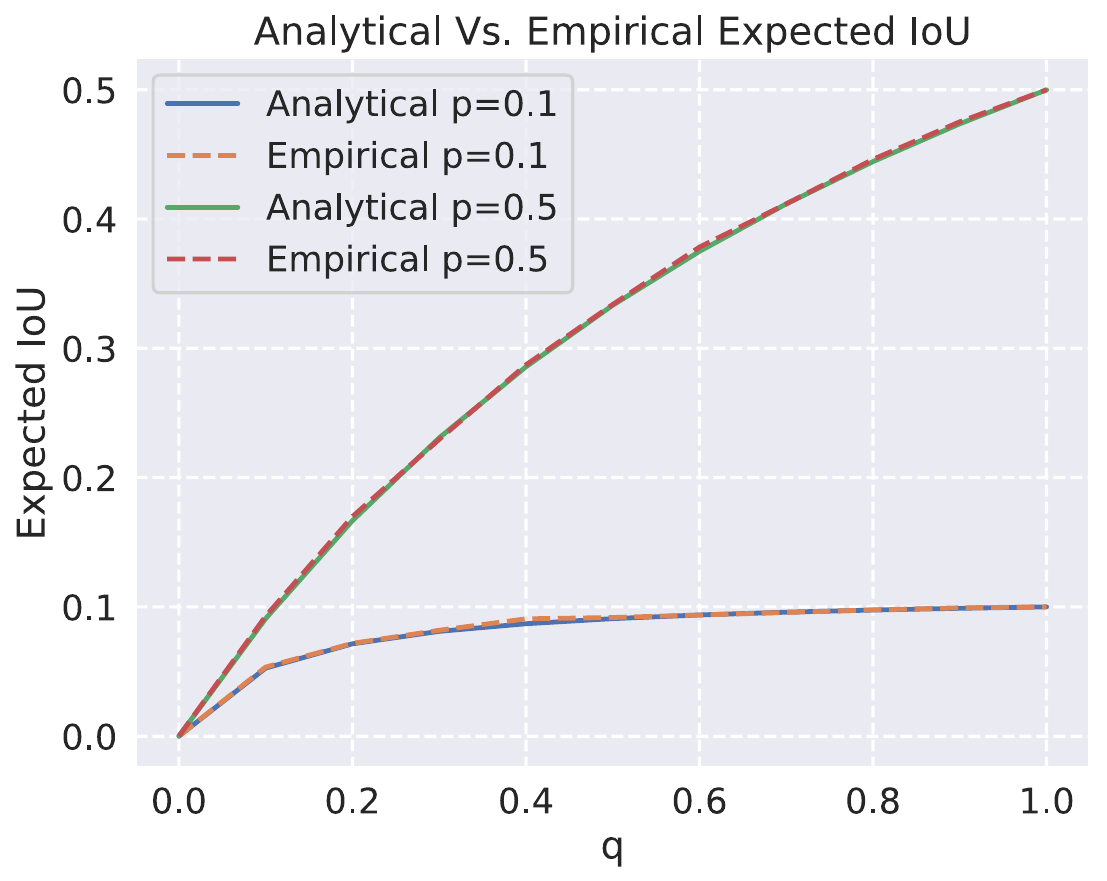}
\caption{Empirical Validation of Theorem~\ref{theorem:iou}} \label{fig:iou_validation}
\end{figure}

\begin{theorem}\label{theorem:iou}
    Consider two SDRs with sparsity defined as random variables $p \sim \mathcal{U}(0,1)$ and $q\sim \mathcal{U}(0,1)$, the expected Jaccard Index (i.e., IoU) of the two random SDRs follows: $$ \frac{pq}{p+q-pq} $$
\end{theorem}

\paragraph{Proof:} Given the sparsity random variables of both SDRs (i.e., $p$ and $q$) and the SDR size $n$, the number of active bits at the same location in both SDRs is equal to the joint probability of both SDRs being active multiplied by the SDR size (i.e., $npq$). The union of both SDRs is the total number of active bits minus the active bits in both SDRs, which is equal to $ np + nq - npq$. Therefore, the expected intersection over union is $\frac{npq}{np+nq-npq} = \frac{pq}{p+q-pq}$

\paragraph{Empirical Validation:} We perform empirical validation of the above theorem as shown in figure~\ref{fig:iou_validation}. The sparsity of the first SDR is fixed at 0.1 and 0.5. We vary the sparsity of the second SDR between 0.0 and 1.0 in steps of 0.1 and calculate the average IoU over a population of 1000 pairs of SDR for every setting. Empirical results agree with the derived formulation in theorem~\ref{theorem:iou}.

\section{Notations}

The notations used in our paper is summarized in Table~\ref{tab:notations}.

\begin{table}[h]
    \centering
    \renewcommand{\arraystretch}{1.25}
    \caption{Table of Notations}
    \label{tab:notations}
    \begin{tabular}{c|p{0.7\columnwidth}}
        \textbf{Symbol} & \textbf{Description} \\
        \specialrule{1.5pt}{0pt}{1.5pt}
        
        $\boldsymbol{A}$  & Learnable transition weight matrix \\
        
        $\boldsymbol{B}$  & Learnable emission weight matrix \\

        $\boldsymbol{x}_t$  & Observation at time $t$ \\

        $\tilde{\boldsymbol{x}}_t$  & Noisy observation at time $t$ during recall \\

        $\boldsymbol{z}_t$  & Posterior Latent state after observing $\boldsymbol{x}_t$ (i.e., single possibility) \\

        $\hat{\boldsymbol{z}}_t$  & Prior Latent state before observing $\boldsymbol{x}_t$ (i.e., multiple possibilities) \\

        $\boldsymbol{a}_t$  & Predicted latent logits at time $t$ (i.e., $\boldsymbol{A} \cdot \boldsymbol{z}_{t-1}$) before applying threshold \\

        $\delta$  & Threshold function: $\mathbb{R}^{N_c \times N_k} \mapsto \{0, 1\}^{N_c \times N_k}$ or $\mathbb{R}^{N_c} \mapsto \{0, 1\}^{N_c}$  \\

        $\eta$ & Hebbian learning strength for adjusting the synaptic weights\\

        $\uparrow$ & Projection operator adds context to a 1d observation state\\

        $\downarrow$ & Projection operator removes context from a 2d latent state\\

        $I$ & Function for computing the indices of active bits in an SDR\\

        $\pi$ & Random permutation function\\

        \midrule

        $N_c$ & Input size of a pattern\\

        $N_k$ & Number of neurons per minicolumn for context encoding\\

        $W$ & Number of active bits in a Sparse Distributed Representation (SDR)\\

        $S$ & Sparsity of SDR, calculated as $W/N$\\

        $T$ & Number of patterns in one sequence (i.e., sequence length)\\

        $d$ & Degree of polynomial in Hopfield separation function\\

        $L$ & Number of Layers used in temporal Predictive Coding (tPC)\\

        \specialrule{1.5pt}{0.5pt}{0pt}        
    \end{tabular}
\end{table}

\section{Datasets}\label{app:datasets}

\paragraph{Synthetic} For synthetic experiments, we generate SDRs with the specified size $N_c$ and uniformly initialized active bits $W$ to match the required sparsity $S$. In many of the experiments, $N_c$ is set to 100 with 5 active bits, unless otherwise specified. For Hopfield experiments, we set the sparsity to 50\% to improve its performance.

\paragraph{Protein Sequences} We use the dataset ProteinNet~7~\cite{alquraishi2019proteinnet} to extract protein sequences. Each sequence consists of a chain of Amino Acids. In the dataset there are only 20 different types of Amino Acids (i.e., vocabulary) creating long protein sequences with hundreds of Amino Acids. The dataset is reported in the fasta format, where each Amino Acid is represented with a single-letter code. We create a dictionary mapping from the Amino Acid types to random SDRs with $N_c=100$ and $W=5$ to train the models. When choosing the sequences, we ensure that the starting Amino Acid is unique for all the dataset sequences to avoid ambiguous predictions in the continual learning evaluation. A sample of the protein sequence is provided below:

\begin{quote}
\begin{lstlisting}[breaklines=true, showstringspaces=false, basicstyle=\small, backgroundcolor=\color{teal!5}]
    MGAAASIQTTVNTLSERISSKLEQEANASAQTKCDIEIGNFYIRQNHGCN
    LTVKNMCSADADAQLDAVLSAATETYSGLTPEQKAYVPAMFTAALNIQTS
    VNTVVRDFENYVKQTCNSSAVVDNKLKIQNVIIDECYGAPGSPTNLEFIN
    TGSSKGNCAIKALMQLTTKATTQIAPKQVAGTGVQFYMIVIGVIILAALF
    MYYAKRMLFTSTNDKIKLILANKENVHWTTYMDTFFRTSPMVIATTDMQN
\end{lstlisting}
\end{quote}

\paragraph{Text} To evaluate the generative ability of PAM, we use a dataset of most frequently used English words. For preprocessing, we extract one hundred 4-letter words from the dataset and create a mapping dictionary from all the unique letters in the dataset to random SDRs with $N_c=100$ and $W=5$ (except for AHN; $W=0.5N_c$). The dataset contains many words with ambiguous future possibilities. The selected words are provided below:

\begin{quote}
\begin{lstlisting}[breaklines=true, showstringspaces=false, basicstyle=\small, backgroundcolor=\color{teal!5}]
    that with they have this from word what some were
    when your said each time will many then them like
    long make look more come most over know than call
    down side been find work part take made live back
    only year came show good give name very just form
    help line turn much mean move same tell does want
    well also play home read hand port even land here
    must high such went kind need near self head page
    grow food four keep last city tree farm hard draw
    left late real life open seem next walk ease both
\end{lstlisting}
\end{quote}

\begin{figure}[h]
\centering
\includegraphics[width = \linewidth]{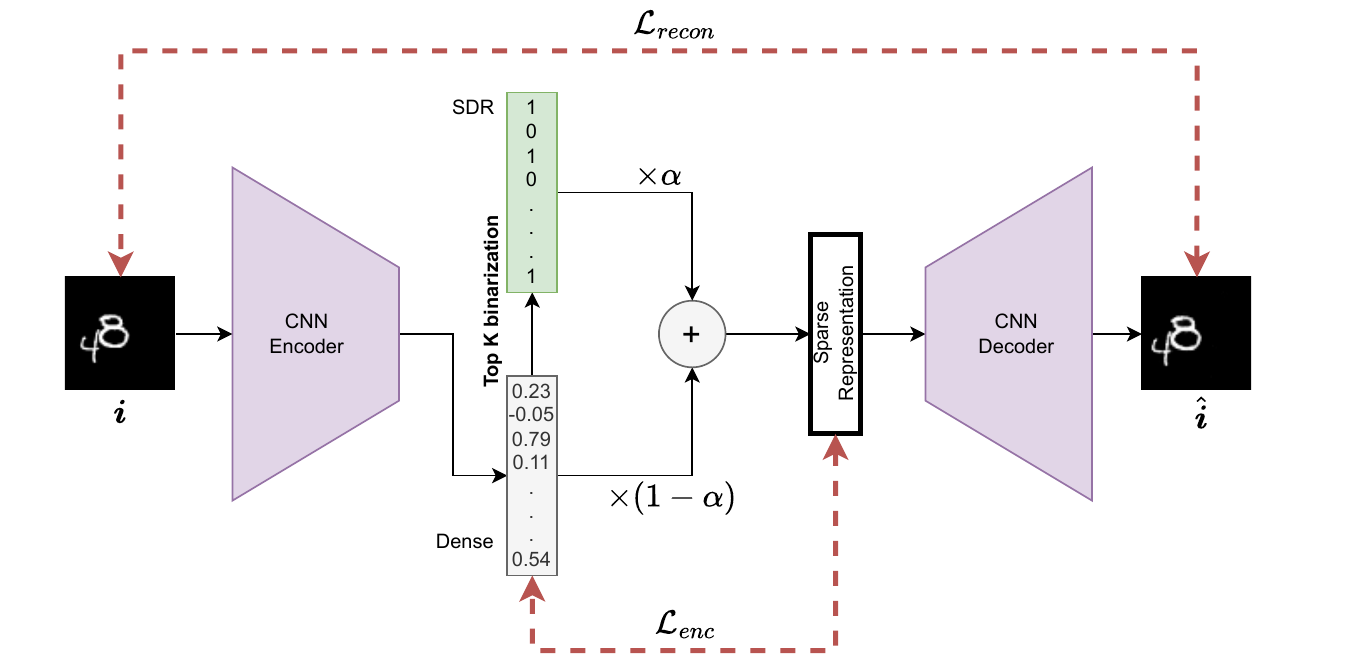}
\caption{Architecture of the SDR Autoencoder} \label{fig:sae}
\end{figure}

\paragraph{Vision} In our experiment, we evaluate on sequences extracted from Moving MNIST~\cite{srivastava2015unsupervised}, CLEVRER~\cite{yi2019clevrer} as well as synthetically generated sequences of CIFAR~\cite{krizhevsky2009learning} images. In order to convert images to SDRs and SDRs back to images while encoding semantics into the SDRs, we design an SDR AutoEncoder. The goal is to force the bottleneck representation of the autoencoder to become a sparse binary representation, where two visually similar images would result in two SDRs with high overlap of active neurons. We simply design a CNN autoencoder with 3-layer CNN encoder and 3-layer CNN decoder, and apply top K binarization operation on the bottleneck embedding during training. The full architecture of the SDR autoencoder is shown in Figure~\ref{fig:sae}.

\begin{figure}[h]
\centering
\includegraphics[width = \linewidth]{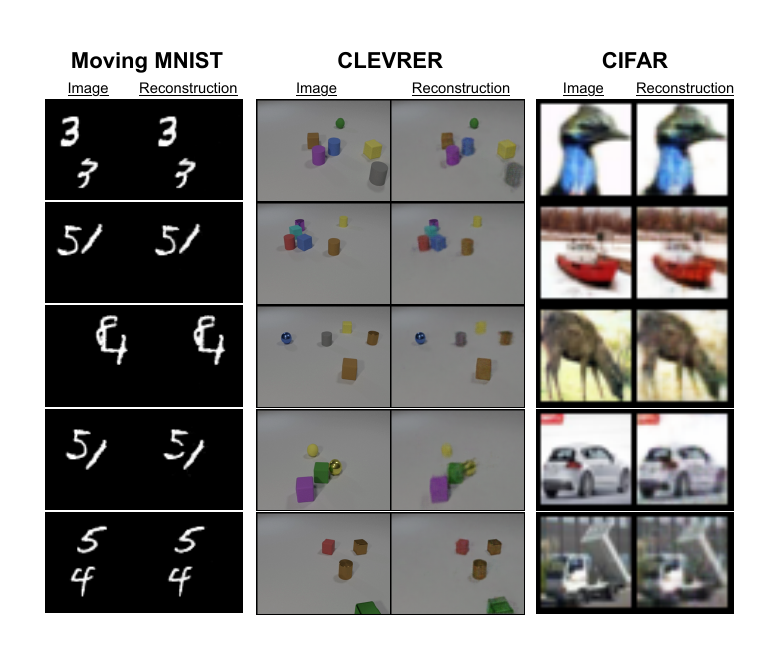}
\caption{Examples of Autoencoder reconstructions from SDRs for all three datasets} \label{fig:sae_examples}
\end{figure}

In practice, we use a weighted average of the SDR and Dense representation to allow gradients of the reconstruction loss to freely propagate into the encoder. The weight of the SDR (i.e., $\alpha$) is gradually and linearly increased (from 0.0 to 1.0) with the number of training epochs. This gradual increase in fundamental to the training of the SDR Autoencoder as it smooths the loss landscape and allows the model to distribute the active bits on the full SDR. The total mse loss becomes $\mathcal{L}_{enc} + \mathcal{L}_{recon}$. We use Adam optimizer with a learning rate of $\num{1e-4}$. For Moving MNIST we use a bottleneck embedding (i.e., $N_c$) of size 100 with 5 active bits, whereas for more complex datasets (i.e., CLEVRER, CIFAR), we use an SDR of size 200 with 10 active bits. We show examples of the autoencoder reconstruction with full binary SDR (i.e., $\alpha=1$) for all three datasets in Figure~\ref{fig:sae_examples}.

\section{Implementation Details}

In this section we describe the implementation details and hyperparameters of each method. For each model, we optimize a single set of hyperparameters for all the experiments.

\paragraph{PAM} The neurons in both, transition and emission, functions are fully connected. We \textit{do not} assume any of the weight matrices are symmetric. All synaptic weights are initialized by sampling from a normal distribution with a mean of 0.0 and a standard deviation of 0.1. All $\eta^+$ values in Equations~\ref{eqn:heb_A}~\&~\ref{eqn:heb_B} are set to $0.1$. $\eta_B^-$ is set to $-0.1$, while $\eta_A^{-}$ is set to $0.0$ to avoid forgetting previous possibilities when learning new transitions; PAM learns a union of possibilities. The threshold for the $\delta$ function is set as a function of the SDR sparsity. For the transition function, we use a threshold of $0.8W$, where $W$ is the active number of bits in the latent state ($\boldsymbol{z}$) SDR. For the emission function, we use a threshold of $0.1W$, where $W$ is the active number of bits in the observation state ($\boldsymbol{x}$) SDR. During offline generation we sample an initial $\Tilde{\boldsymbol{x}}$ from $\downarrow \hat{\boldsymbol{z}}$ with $W=1$ active neurons. During generation, we set the maximum number of attractor iterations to 100, but stop iterating when the energy of the state converges to a local minimum. During sequence learning, we update $\boldsymbol{A}$ and $\boldsymbol{B}$ iteratively until the transition is learned, before learning the next transition. This iterative weight update makes the model insensitive to the hyperparameter values $\eta$. Both $\boldsymbol{A}$ and $\boldsymbol{B}$ are always clamped in the range $[-1, 1]$. The states $\boldsymbol{z}$ are flattened into a single dimension before applying the learning rule in Equation~\ref{eqn:heb_A}. Binary representations (i.e., \{0, 1\}) are used as inputs.

\paragraph{Temporal Predictive Coding} For the tPC architecture, we use learning rate of 1e-4 for 800 learning iterations. When a 2-layer tPC model is used, the inference learning rate is set to 1e-2 for 400 inference iterations. Also, the hidden size is set to twice the input size. We found that these parameters work best for all of the experiments and allow the model to fully converge. Bipolar representations (i.e., \{-1, 1\}) are used as inputs.

\paragraph{Asymmetric Hopfield Network} The Hopfield model does not require hyperparameters other than the ablated separation function. In many experiments, we use a polynomial separation function with degree $d$ set to 1 or 2. Bipolar representations (i.e., \{-1, 1\}) are used as inputs.

\section{Experiments}\label{app:experiments}

In this section we describe the setup of each figure in the main paper and provide additional quantitative and qualitative results for each task. \textit{All experiments are run for 10 different seeds/trials. We report the mean and standard deviation in all the figures and tables.}

\subsection{Sequence Capacity}

\paragraph{Figure~\ref{fig:results1}~A} This experiment plots the maximum offline sequence length (i.e., sequence capacity, $T_{max}$) at different input sizes. The input size $N_c$ is varied from $10$ to $100$ while the number of active bits $W$ is fixed to $5$. We compare variants of our model with $N_k$ set to $4$ and $8$ to temporal predictive coding (tPC) and Asymmetric Hopfield Network (AHN). We observe that AHN completely fails as the sparsity $S$ of the pattern decreases, therefore we also compare to AHN with the sparsity set to 50\% (i.e., $W=0.5N_c$). For AHN models, we experiment with a polynomial exponential function with degree $1$ and $2$, as recently proposed~\cite{chaudhry2024long} and used for evaluation in recent papers~\cite{tang2024sequential}. All models in this experiment are set to recall/generate in an offline manner, where only the first input is provided. PAM outperforms all other methods and has the potential to improve further by expanding the context neurons $N_k$. The patterns in this experiment are uncorrelated such that each pattern has active bits uniformly initialized.

\paragraph{Figure~\ref{fig:results1}~B} This experiment plots the effect of sequence correlation on the maximum offline capacity. The higher the correlation value, the more exact repetitions of patterns are available in the sequence. We enforce correlation by limiting the number of unique patterns (i.e., vocab) used to create the sequence. All patterns in this experiment are set to a size of $N_c=100$ and $W=5$ (except for AHN which is set at $W=0.5N_c$). Results show that the capacity of all other methods sharply drops when correlation is introduced. PAM retains most of its original capacity.

\paragraph{Figure~\ref{fig:results1}~E \& F} In this experiment, we provide a qualitative example of a short sequence ($T=10$) with high correlation ($0.8$). The sequence is learned by all the methods, then we perform offline (\textbf{E}) and online (\textbf{F}) recall on the sequence. We use the SDR autoencoder to create SDRs from these CIFAR images for training and recall. The SDRs have a size $N_c$ of 200 and $W=10$. In the offline recall, only the first input is provided and the model auto-regressively generates the full sequence using its own predictions at every time step. In online recall, the models perform a single step prediction and always uses the groundtruth input at every time step to perform predictions. Results show that only PAM can retain a context of correlated sequence and accurately predicts into the future based on this context.

\begin{figure}[H]
\centering
\includegraphics[width = \linewidth]{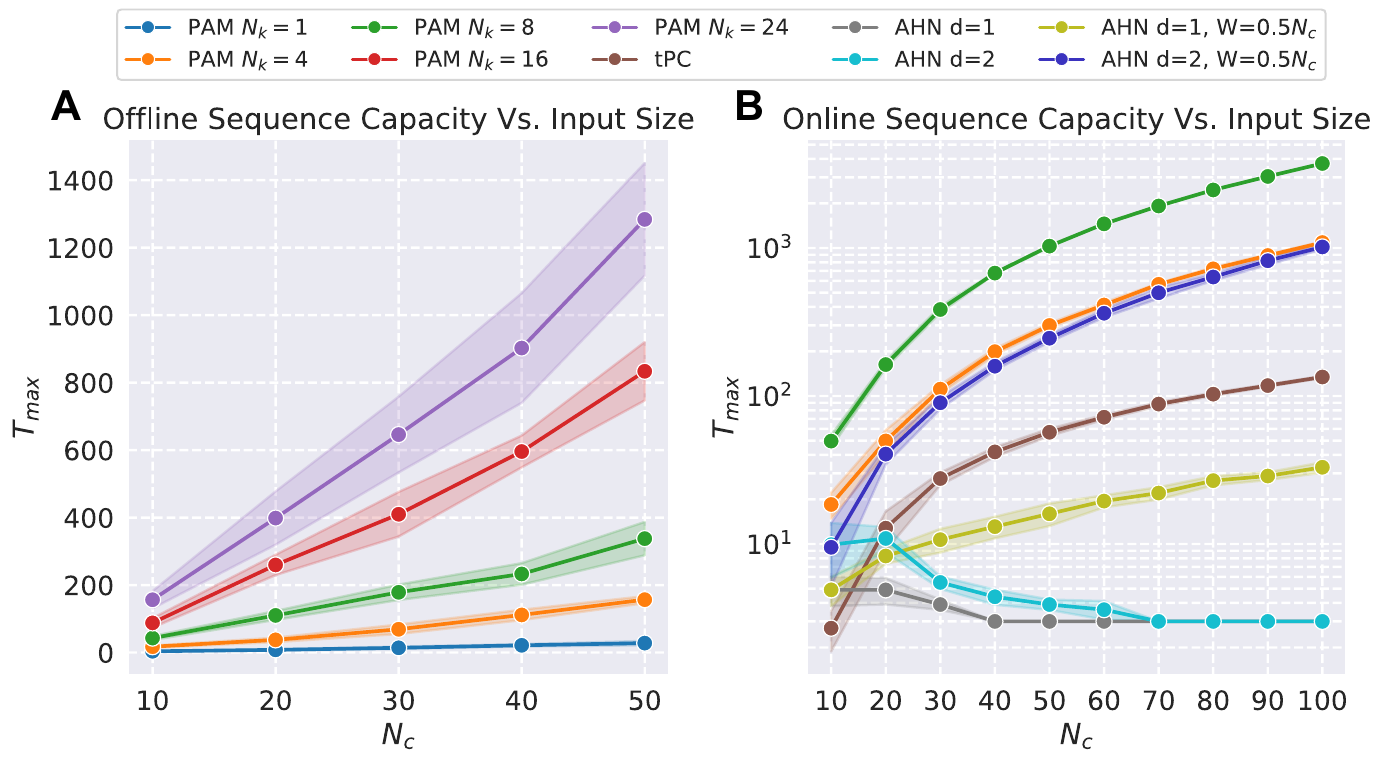}
\caption{Additional sequence capacity experiments. \textbf{A}: scaling of the offline sequence capacity with context memory size $N_k$ and input size $N_c$. \textbf{B}: Online sequence capacity of various methods.} \label{fig:app_capacity_scale}
\end{figure}

\paragraph{Figure~\ref{fig:app_capacity_scale}~A} In this experiment, we show the effect of scaling the model context memory beyond a simple $N_k=4$ and $N_k=8$. We show that when using $N_k=16$ and $N_k=24$, PAM can model much longer sequences. We vary the input size $N_c$ from 10 to 50 and report the offline sequence capacity of the model as ablations.

\paragraph{Figure~\ref{fig:app_capacity_scale}~B} Similar to the experiment plotted in Figure~\ref{fig:results1}~\textbf{A}, we report the sequence capacity with input size $N_c$. However, this experiment evaluates the online generation capacity, where the model uses the correct pattern at every prediction time step instead of using its own prediction from the previous time step. Results show that PAM significantly increased in capacity (three times in some cases), whereas the other methods have not increased as much in modeling longer sequences.

\begin{figure}[H]
\centering
\includegraphics[width = \linewidth]{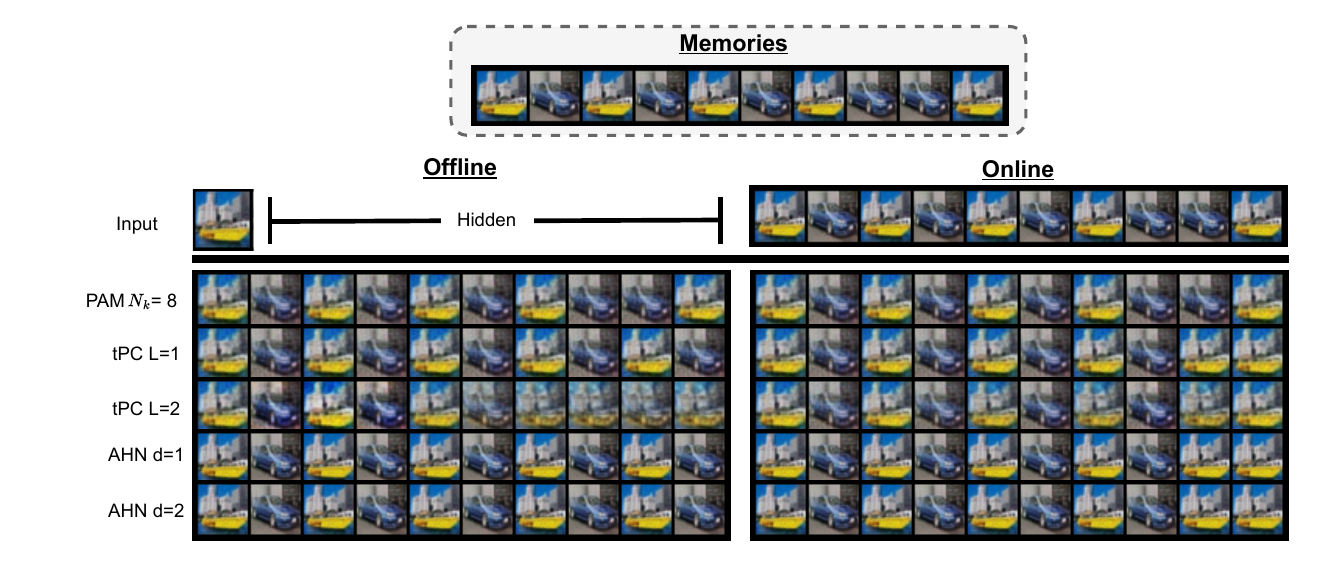}
\caption{Additional qualitative example of correlated sequential memory with CIFAR images.} \label{fig:app_capacity_qual}
\end{figure}

\paragraph{Figure~\ref{fig:app_capacity_qual}} We provide additional qualitative example on a different highly correlated sequence. The result shows a different failure mode for AHN, whereas PAM still performs well.

\subsection{Catastrophic Forgetting}

\paragraph{Figure~\ref{fig:results2}~A} We benchmark the performance of difference models in the challenging continual learning setup. The models are expected to avoid catastrophic forgetting by not overwriting previously learned sequences. In this experiment, we use 50 sequences, each with size $N_c=100$ and length $T=10$. We vary the correlation of the sequences from $0.0$ to $0.5$ and compute the backward transfer metric with the normalized IoU as the measure of similarity. Results show that AHN can avoid catastrophic forgetting when the sequence are uncorrelated, but quickly drops in performance with correlation. tPC fails in retaining learned sequences regardless of correlation. PAM performs well with more context neurons $N_k$. When setting $N_k$ to $1$, the model fails to retain its knowledge due to the decreased context modeling capability with a single context neuron.

\paragraph{Figure~\ref{fig:results2}~B} In this experiment, we report the performance of the models on protein sequences. This is a more challenging setup due to the long sequence (few hundreds on average) with high correlation (only 20 unique Amino Acids). We show a similar trend, where the other methods fail due to high correlation or sequence lengths. PAM outperforms the other methods when using context memory $N_k$ of $16$ or $24$. All Amino Acids types are converted to fixed and randomly initialized SDRs with $N_c=100$. The sparsity is set similar to sequence capacity experiments (i.e., $W=5$ and $W=0.5N_c$).

\paragraph{Figure~\ref{fig:results2}~F} We provide qualitative results on a simple experiment with 2 sequences from moving MNIST. The models learn the first sequence then learn the second sequence. The models are not allowed to train on the first sequence after they have trained on the second sequence. We then perform online generation on the first sequence with all models. We use the SDR autoencoder to generate SDRs for all images in the sequences, the SDRs have $N_c=100$ with $W=5$ (for all methods except AHN). The results show that PAM can recall the full sequence even after being trained on another sequence. Other methods fail in this simple task even in online recall setup.

\begin{figure}[H]
\centering
\includegraphics[width = \linewidth]{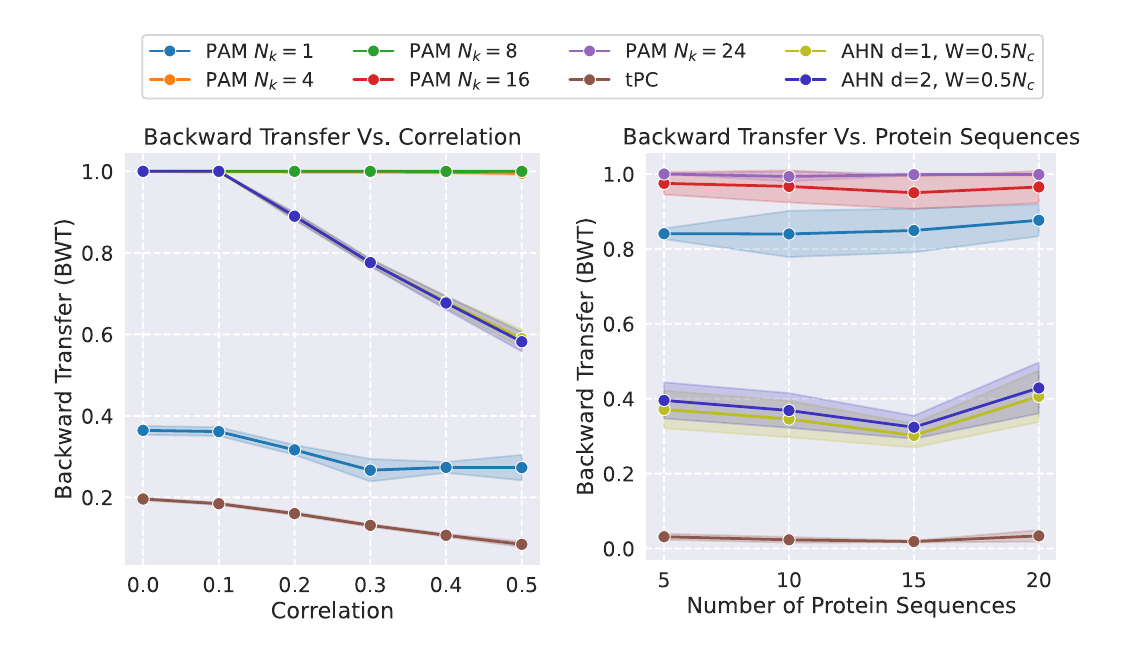}
\caption{Catastrophic forgetting experiments in online manner. Results are shown on synthetic dataset of SDR sequences with different correlations and on protein sequences.} \label{fig:app_online_forgetting}
\end{figure}

\paragraph{Figure~\ref{fig:app_online_forgetting}} We provide continual learning results similar to Figures~\ref{fig:results2}~\textbf{A}~\&~\textbf{B}; however, instead of offline generation, we perform the evaluation in onine manner.

\begin{figure}[H]
\centering
\includegraphics[width = \linewidth]{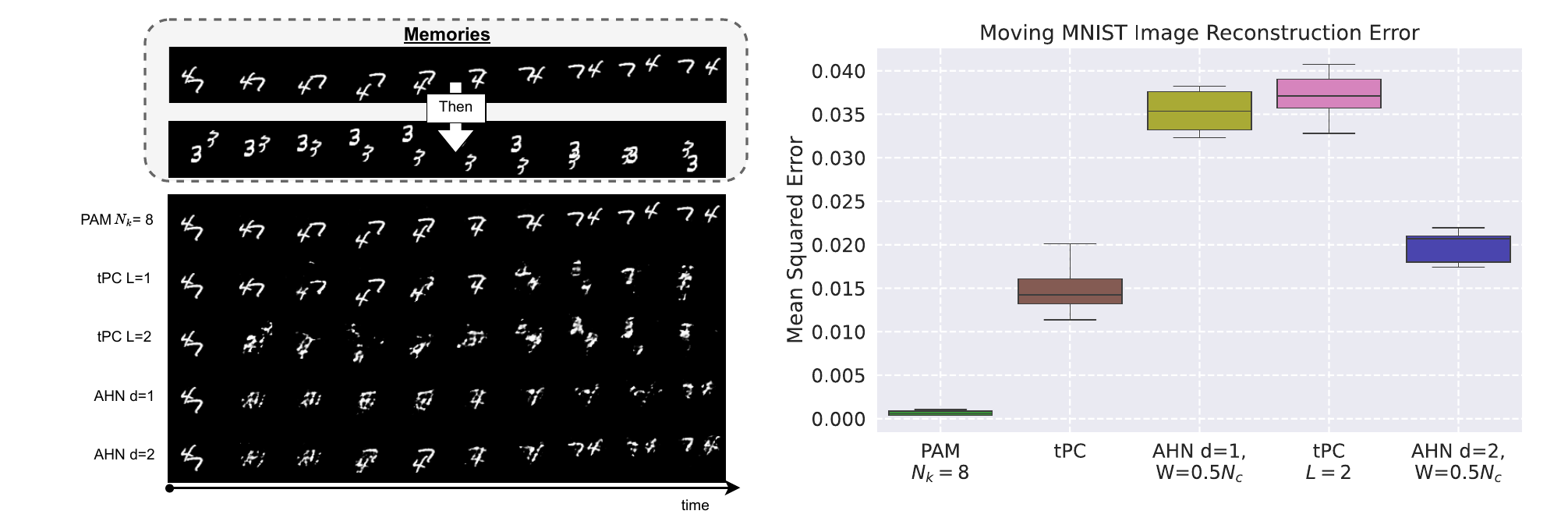}
\caption{Additional qualitative catastrophic forgetting visualization on Moving MNIST and quantitative results on the reconstruction error of 10 Moving MNIST random examples} \label{fig:app_mmnist_forgetting}
\end{figure}

\paragraph{Figure~\ref{fig:app_mmnist_forgetting}} We provide additional qualitative results on Moving MNIST, as well as quantitative results averaged over 10 trials of MNIST sequence pairs. These quantiative results are reported as the Mean Squared Error of the reconstructed image. Results show that PAM reports the lowest error with a much smaller variance.

\paragraph{Tables~\ref{tbl:bwt_pam1},~\ref{tbl:bwt_pam4},~\ref{tbl:bwt_tpc},~\&~\ref{tbl:bwt_ahn}} The Backward Transfer metric (BWT) to evaluate catastrophic forgetting is computed by taking the average of the performance on previously learned sequences after training on a new sequence. In addition to plotting this average in previous experiments, we provide the full tables for one of experiment as an example. All tables show results on 10 sequences and $N_c=100$. The BWT metric is calculated as the average of the similarity metric reported in these tables.

\begin{table}[H]
    \centering
    \caption{Catastophic forgetting experiment results on 10 sequences for PAM with $N_k=1$. The table shows the mean normalized IoU and standard deviation of previous learned sequences after training on new sequences. The Backward Transfer metric is the average of all the shown numbers. Results are averaged over 10 trials.}\label{tbl:bwt_pam1}
    \resizebox{\linewidth}{!}{%
    \begin{tabular}{|cc|cccccccccc|}
        \toprule
        \multicolumn{2}{|c|}{\multirow{2}{*}{\diagbox[width=2cm]{Train}{Test}}} & \multicolumn{10}{c|}{Sequence ID} \\
        & & 1 & 2 & 3 & 4 & 5 & 6 & 7 & 8 & 9 & 10\\
        \midrule
        \multirow{5}{*}{
    \rotatebox[origin=l]{90}{Sequence ID   \quad\quad\; }} 
                    & 1 & - & - & - & - & - & - & - & - & - & - \\
                    & 2 & 0.692 ± 0.220 & - & - & - & - & - & - & - & - & - \\
                    & 3 & 0.654 ± 0.252 & 0.586 ± 0.314 & - & - & - & - & - & - & - & - \\
                    & 4 & 0.581 ± 0.200 & 0.619 ± 0.411 & 0.759 ± 0.229 & - & - & - & - & - & - & - \\
                    & 5 & 0.634 ± 0.236 & 0.553 ± 0.348 & 0.817 ± 0.200 & 0.823 ± 0.215 & - & - & - & - & - & - \\
                    & 6 & 0.617 ± 0.224 & 0.501 ± 0.322 & 0.695 ± 0.289 & 0.546 ± 0.329 & 0.637 ± 0.299 & - & - & - & - & - \\
                    & 7 & 0.660 ± 0.246 & 0.545 ± 0.343 & 0.706 ± 0.265 & 0.490 ± 0.239 & 0.604 ± 0.325 & 0.716 ± 0.293 & - & - & - & - \\
                    & 8 & 0.631 ± 0.231 & 0.657 ± 0.317 & 0.693 ± 0.292 & 0.527 ± 0.236 & 0.633 ± 0.300 & 0.571 ± 0.297 & 0.748 ± 0.268 & - & - & - \\
                    & 9 & 0.700 ± 0.237 & 0.531 ± 0.357 & 0.723 ± 0.294 & 0.645 ± 0.286 & 0.584 ± 0.277 & 0.605 ± 0.284 & 0.613 ± 0.265 & 0.539 ± 0.267 & - & - \\
                    & 10 & 0.659 ± 0.235 & 0.488 ± 0.317 & 0.602 ± 0.326 & 0.498 ± 0.260 & 0.532 ± 0.246 & 0.630 ± 0.257 & 0.664 ± 0.327 & 0.660 ± 0.330 & 0.575 ± 0.313 & - \\
        \bottomrule
    \end{tabular}
    }
\end{table}

\begin{table}[H]
    \centering
    \caption{Catastophic forgetting experiment results on 10 sequences for PAM with $N_k=4$. The table shows the mean normalized IoU and standard deviation of previous learned sequences after training on new sequences. The Backward Transfer metric is the average of all the shown numbers. Results are averaged over 10 trials.}\label{tbl:bwt_pam4}
    \resizebox{\linewidth}{!}{%
    \begin{tabular}{|cc|cccccccccc|}
        \toprule
        \multicolumn{2}{|c|}{\multirow{2}{*}{\diagbox[width=2cm]{Train}{Test}}} & \multicolumn{10}{c|}{Sequence ID} \\
        & & 1 & 2 & 3 & 4 & 5 & 6 & 7 & 8 & 9 & 10\\
        \midrule
        \multirow{5}{*}{
    \rotatebox[origin=l]{90}{Sequence ID   \quad\quad\; }} 
                    & 1 & - & - & - & - & - & - & - & - & - & - \\
                    & 2 & 1.000 ± 0.000 & - & - & - & - & - & - & - & - & - \\
                    & 3 & 1.000 ± 0.000 & 1.000 ± 0.000 & - & - & - & - & - & - & - & - \\
                    & 4 & 1.000 ± 0.000 & 1.000 ± 0.000 & 1.000 ± 0.000 & - & - & - & - & - & - & - \\
                    & 5 & 1.000 ± 0.000 & 1.000 ± 0.000 & 1.000 ± 0.000 & 1.000 ± 0.000 & - & - & - & - & - & - \\
                    & 6 & 1.000 ± 0.000 & 1.000 ± 0.000 & 1.000 ± 0.000 & 1.000 ± 0.000 & 1.000 ± 0.000 & - & - & - & - & - \\
                    & 7 & 1.000 ± 0.000 & 1.000 ± 0.000 & 1.000 ± 0.000 & 1.000 ± 0.000 & 1.000 ± 0.000 & 1.000 ± 0.000 & - & - & - & - \\
                    & 8 & 1.000 ± 0.000 & 1.000 ± 0.000 & 1.000 ± 0.000 & 1.000 ± 0.000 & 1.000 ± 0.000 & 1.000 ± 0.000 & 1.000 ± 0.000 & - & - & - \\
                    & 9 & 1.000 ± 0.000 & 1.000 ± 0.000 & 1.000 ± 0.000 & 1.000 ± 0.000 & 1.000 ± 0.000 & 1.000 ± 0.000 & 1.000 ± 0.000 & 1.000 ± 0.000 & - & - \\
                    & 10 & 1.000 ± 0.000 & 1.000 ± 0.000 & 1.000 ± 0.000 & 1.000 ± 0.000 & 1.000 ± 0.000 & 1.000 ± 0.000 & 1.000 ± 0.000 & 1.000 ± 0.000 & 1.000 ± 0.000 & - \\
        \bottomrule
    \end{tabular}
    }
\end{table}

\begin{table}[H]
    \centering
    \caption{Catastophic forgetting experiment results on 10 sequences for tPC. The table shows the mean normalized IoU and standard deviation of previous learned sequences after training on new sequences. The Backward Transfer metric is the average of all the shown numbers. Results are averaged over 10 trials.}\label{tbl:bwt_tpc}
    \resizebox{\linewidth}{!}{%
    \begin{tabular}{|cc|cccccccccc|}
        \toprule
        \multicolumn{2}{|c|}{\multirow{2}{*}{\diagbox[width=2cm]{Train}{Test}}} & \multicolumn{10}{c|}{Sequence ID} \\
        & & 1 & 2 & 3 & 4 & 5 & 6 & 7 & 8 & 9 & 10\\
        \midrule
        \multirow{5}{*}{
    \rotatebox[origin=l]{90}{Sequence ID   \quad\quad\; }} 
                    & 1 & - & - & - & - & - & - & - & - & - & - \\
                    & 2 & 0.452 ± 0.230 & - & - & - & - & - & - & - & - & - \\
                    & 3 & 0.180 ± 0.135 & 0.360 ± 0.270 & - & - & - & - & - & - & - & - \\
                    & 4 & 0.148 ± 0.102 & 0.326 ± 0.257 & 0.462 ± 0.311 & - & - & - & - & - & - & - \\
                    & 5 & 0.095 ± 0.047 & 0.148 ± 0.040 & 0.253 ± 0.166 & 0.393 ± 0.339 & - & - & - & - & - & - \\
                    & 6 & 0.055 ± 0.038 & 0.102 ± 0.052 & 0.213 ± 0.245 & 0.211 ± 0.199 & 0.344 ± 0.256 & - & - & - & - & - \\
                    & 7 & 0.068 ± 0.037 & 0.062 ± 0.050 & 0.102 ± 0.067 & 0.103 ± 0.087 & 0.215 ± 0.197 & 0.383 ± 0.189 & - & - & - & - \\
                    & 8 & 0.038 ± 0.035 & 0.041 ± 0.045 & 0.052 ± 0.034 & 0.080 ± 0.074 & 0.073 ± 0.062 & 0.232 ± 0.138 & 0.461 ± 0.326 & - & - & - \\
                    & 9 & 0.021 ± 0.022 & 0.032 ± 0.022 & 0.056 ± 0.042 & 0.057 ± 0.027 & 0.079 ± 0.086 & 0.170 ± 0.129 & 0.290 ± 0.159 & 0.284 ± 0.169 & - & - \\
                    & 10 & 0.016 ± 0.019 & 0.023 ± 0.028 & 0.039 ± 0.028 & 0.032 ± 0.039 & 0.058 ± 0.058 & 0.133 ± 0.059 & 0.187 ± 0.134 & 0.288 ± 0.302 & 0.261 ± 0.237 & - \\
        \bottomrule
    \end{tabular}
    }
\end{table}

\begin{table}[H]
    \centering
    \caption{Catastophic forgetting experiment results on 10 sequences for AHN with $d=2$ and $W=0.5N_c$. The table shows the mean normalized IoU and standard deviation of previous learned sequences after training on new sequences. The Backward Transfer metric is the average of all the shown numbers. Results are averaged over 10 trials.}\label{tbl:bwt_ahn}
    \resizebox{\linewidth}{!}{%
    \begin{tabular}{|cc|cccccccccc|}
        \toprule
        \multicolumn{2}{|c|}{\multirow{2}{*}{\diagbox[width=2cm]{Train}{Test}}} & \multicolumn{10}{c|}{Sequence ID} \\
        & & 1 & 2 & 3 & 4 & 5 & 6 & 7 & 8 & 9 & 10\\
        \midrule
        \multirow{5}{*}{
    \rotatebox[origin=l]{90}{Sequence ID   \quad\quad\; }} 
                    & 1 & - & - & - & - & - & - & - & - & - & - \\
                    & 2 & 0.689 ± 0.192 & - & - & - & - & - & - & - & - & - \\
                    & 3 & 0.689 ± 0.192 & 0.595 ± 0.334 & - & - & - & - & - & - & - & - \\
                    & 4 & 0.689 ± 0.192 & 0.595 ± 0.334 & 0.765 ± 0.238 & - & - & - & - & - & - & - \\
                    & 5 & 0.689 ± 0.192 & 0.595 ± 0.334 & 0.765 ± 0.238 & 0.780 ± 0.226 & - & - & - & - & - & - \\
                    & 6 & 0.689 ± 0.192 & 0.595 ± 0.334 & 0.765 ± 0.238 & 0.780 ± 0.226 & 0.692 ± 0.268 & - & - & - & - & - \\
                    & 7 & 0.689 ± 0.192 & 0.595 ± 0.334 & 0.765 ± 0.238 & 0.780 ± 0.226 & 0.692 ± 0.268 & 0.667 ± 0.226 & - & - & - & - \\
                    & 8 & 0.689 ± 0.192 & 0.595 ± 0.334 & 0.765 ± 0.238 & 0.780 ± 0.226 & 0.692 ± 0.268 & 0.667 ± 0.226 & 0.734 ± 0.267 & - & - & - \\
                    & 9 & 0.689 ± 0.192 & 0.595 ± 0.334 & 0.765 ± 0.238 & 0.780 ± 0.226 & 0.692 ± 0.268 & 0.667 ± 0.226 & 0.734 ± 0.267 & 0.558 ± 0.257 & - & - \\
                    & 10 & 0.689 ± 0.192 & 0.595 ± 0.334 & 0.765 ± 0.238 & 0.780 ± 0.226 & 0.692 ± 0.268 & 0.667 ± 0.226 & 0.734 ± 0.267 & 0.558 ± 0.257 & 0.659 ± 0.293 & - \\
        \bottomrule
    \end{tabular}
    }
\end{table}

\subsection{Multiple Possibilities Generation}

In this task, we evaluate the models' ability to generate meaningful sequences and recall the full dataset despite presented with multiple valid possibilities. Ideally, the models are expected to sample a single possibility if trained on sequences with ambiguous future continuations (equally valid possibilities). This is a challenging task for most biologically plausible (e.g., tPC, AHN, etc.) and implausible (e.g., transformers, RNNs, etc) models. Most approaches assume the existence of a full set of possibilities and transform the task from regression to classification (e.g., LLM). For vision tasks, some methods (VQ-VAE and its variants) cluster the dense representations to create this set of possibilities and perform classification. We \textit{do not} assume the existence of a full set of possibilities, but instead perform a true generative evaluation as a regression task.

\paragraph{Figure~\ref{fig:results2}~C} In this experiment, we compute the average normalized IoU of the generated words. The models ability to generate a full sequence with high IoU means it can produce sharp single predictions despite being trained on multiple equally valid future predictions. As the number of words increase, the performance of other models decrease as they struggle to model ambiguous future predictions; however, PAM outperforms the other approaches by sampling from these possibilities.

\paragraph{Figure~\ref{fig:results2}~D} This experiment evaluates the ability of the models to recall the dataset words. We compute the recall as the number of valid unique words generated divided by the total number of words in the dataset. Since PAM is a generative stochastic model, the recall increases with every generation. The other methods are deterministic, therefore do not report an increase in dataset recall with more generations. The other methods completely fail in generating any meaningful words. we use an average IoU threshold of 0.9 to classify a generated word as correct, similar to sequence capacity experiments.

\begin{figure}[H]
\centering
\includegraphics[width = \linewidth]{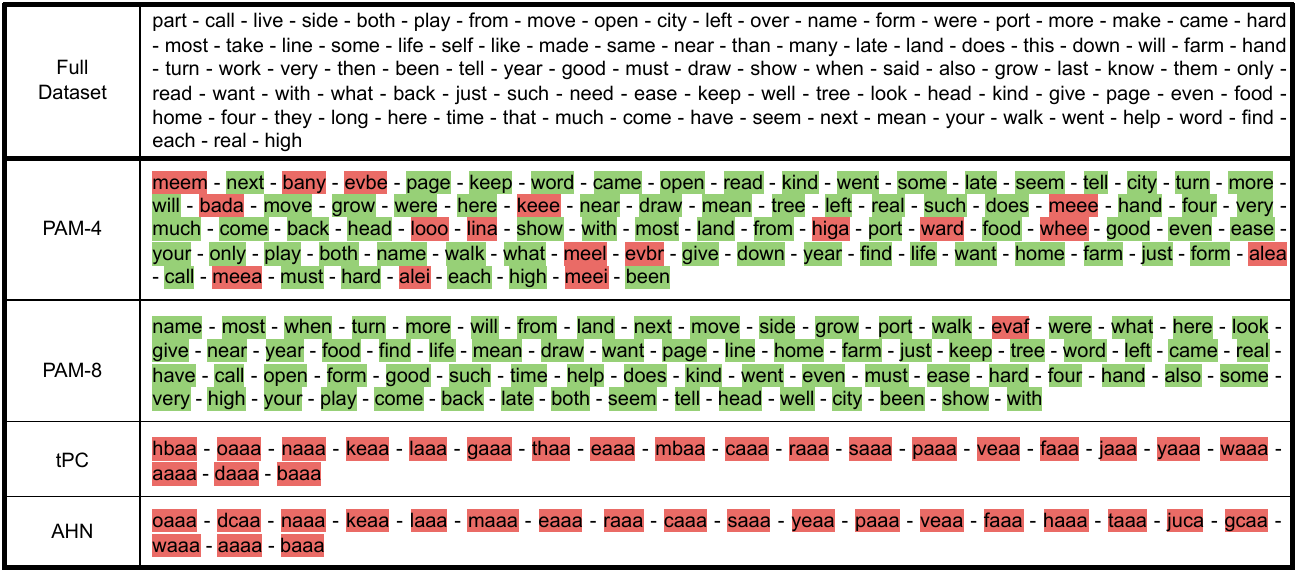}
\caption{Qualitative results showing the generated words from PAM, tPC~\cite{tang2024sequential} and AHN~\cite{chaudhry2024long} Words highlighted in green are available in the dataset (i.e., True positives). Words highlighted in red are not available in the dataset (i.e., False positives).} \label{fig:app_words}
\end{figure}

\paragraph{Figure~\ref{fig:app_words}} We provide qualitative results by showing the unique generated words by different models after 5 dataset generations. PAM $N_k=4$ generates some of the dataset words, but also generates many wrong words. By increasing the context memory neurons to $N_k=8$, the model generates many more correct words and reduces the false positives. The other methods cannot generate meaningful words.

\subsection{Noise Robustness}

\paragraph{Figure~\ref{fig:results1}~C} To evaluate for noise robustness, we plot each model's performance (Normalized IoU) with varying levels of noise added in an online generation setting. The noisy inputs are created by changing a percentage of the active neurons to different uniformly chosen neurons in the SDR. The noise is computed as a percentage for a fair comparison across different SDR sparsities (e.g., tPC vs. AHN). We show the results for sequences of lengths $T=200$ and no correlation. PAM has the ability to compare the noisy input representation to the learned attractors to recover the correct clean input. Therefore, even when the SDR is completely changed, PAM relies on its predictions and completely ignores the noisy input. During generation, PAM always generates (or corrects a noisy input) from within the predicted set of possibilities. The other approaches use the noisy inputs during recall which affects their performance.

\paragraph{Figure~\ref{fig:results2}~E} We provide qualitative results on the CLEVRER dataset. The memories sequence is learned by all the models, then a noisy sequence is used during generation. We only add noise starting from the second pattern in the input sequence. The results show tPC models performing relatively well, yet still outperformed by PAM $N_k=8$. We set $N_c=200$ in the SDR autoencoder to learn the SDRs used in this experiment. We use 40\% noise in this experiment.

\begin{figure}[H]
\centering
\includegraphics[width = \linewidth]{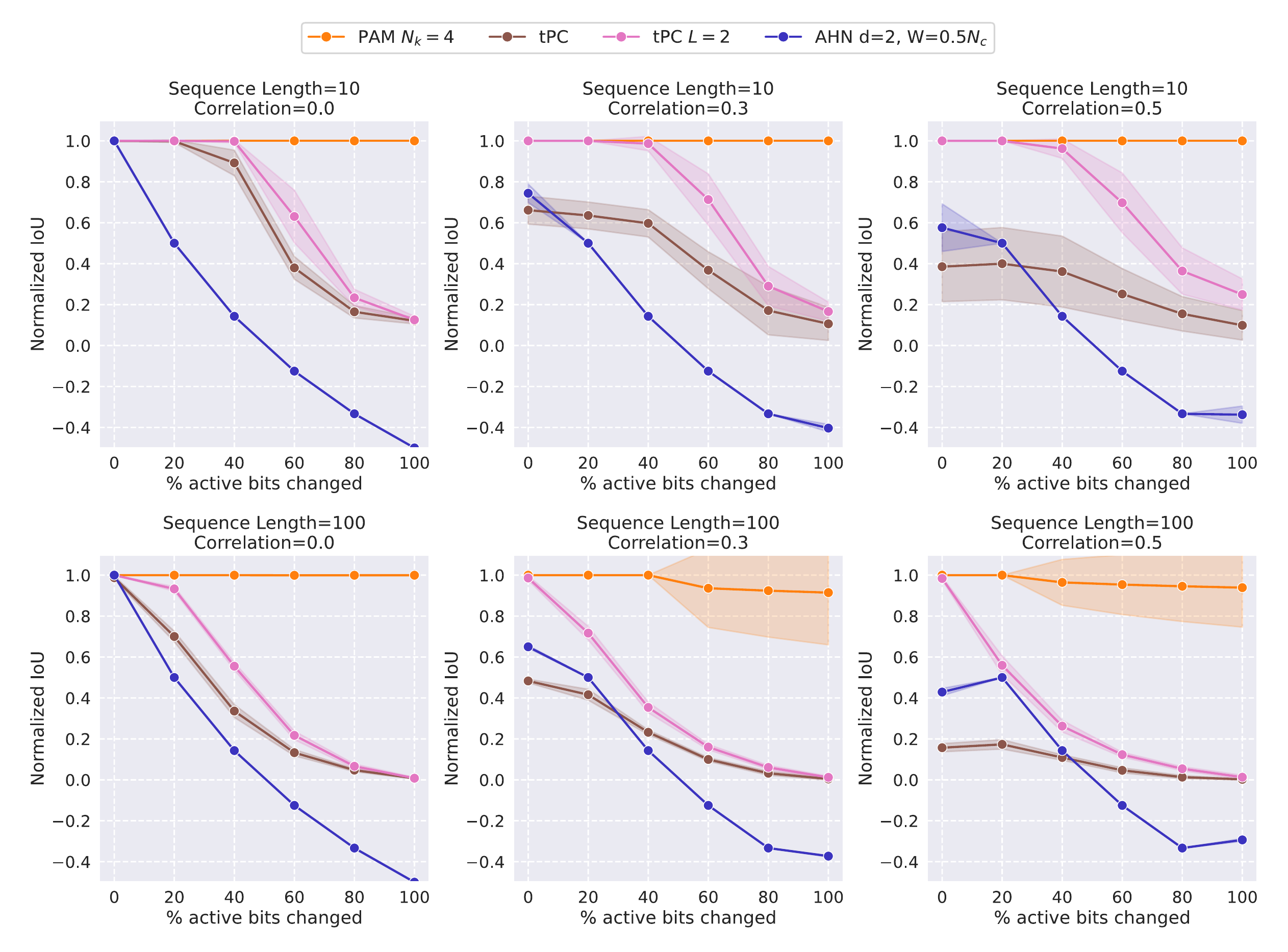}
\caption{The effect of noise on online generation with varying sequence lengths and sequence correlations.} \label{fig:app_sdr_noise}
\end{figure}

\paragraph{Figure~\ref{fig:app_sdr_noise}} We perform additional experiments on varying the sequence lengths and the correlation in the sequence, all the other settings remain the same as in the experiment of Figure~\ref{fig:results1}~\textbf{C}. The results show that with shorter sequences ($\leq 200$), no noise and no correlation, all the models recall the learned sequence well. When higher correlation is used, 2-layered tPC performs relatively well with short sequences (i.e., $T=10$), but fails with longer sequences (i.e., $T=100$). The hopfield model fails more with correlation than sequence length. The added noise affects all reported methods except for PAM, due to its ability to rely on its predictions and attractors to clean the noisy signal.

\begin{figure}[H]
\centering
\includegraphics[width = \linewidth]{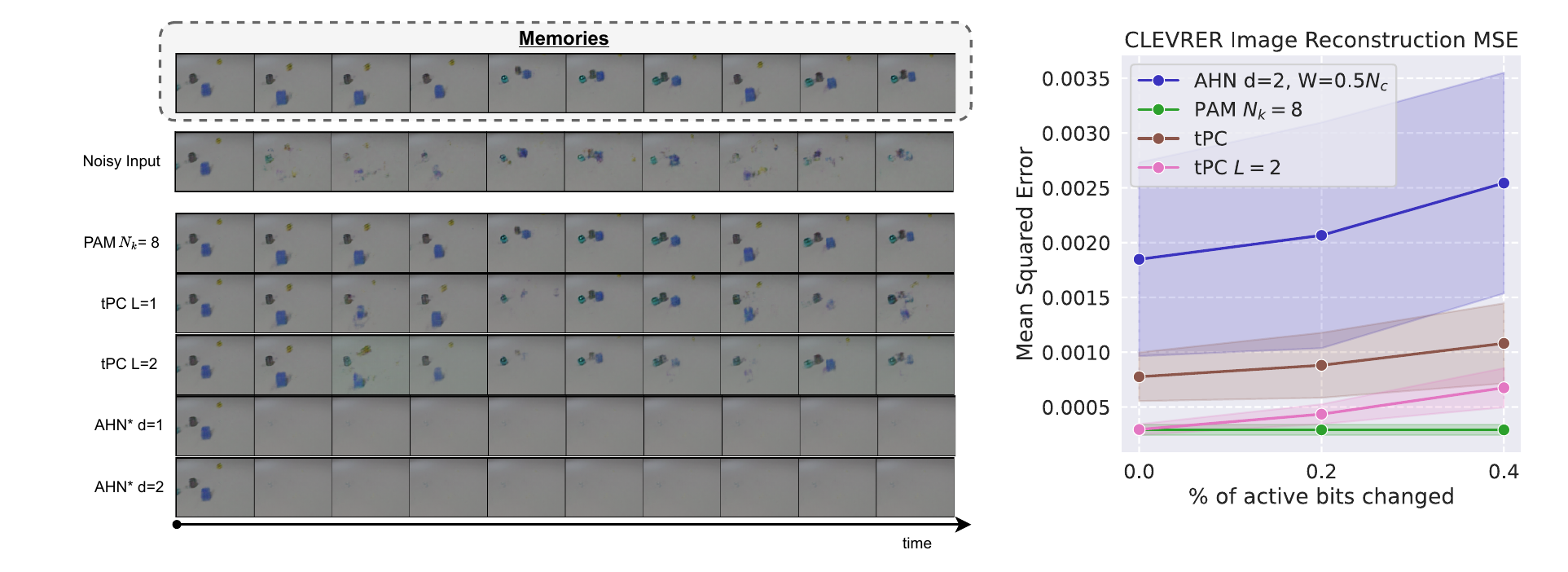}
\caption{Additional qualitative example of online generation with noise on CLEVRER dataset, and quantitative results of reconstruction error over 10 CLEVRER examples at different noise levels.} \label{fig:app_clevrer_noise}
\end{figure}

\paragraph{Figure~\ref{fig:app_clevrer_noise}} We provide an additional qualitative example with similar trend to Figure~\ref{fig:results2}~\textbf{E}. We also provide quantitative results of CLEVRER averaged over 10 experiments. The mean squared error of the generated sequence for multiple models at different noise levels is reported. It is clear that PAM outperforms all methods, and a 2-layered tPC is the second best.

\subsection{Efficiency}

\paragraph{Figure~\ref{fig:results1}~D} We compare the efficiency of the models and show that PAM is at least two order of magnitude more efficient than tPC with 2 layers. A single layer tPC is almost equivalent to PAM with high context memory of $N_k=24$. AHN is highly efficient as the model is not usually trained, but the recall equation is used instead. Therefore, we exclude AHN from the comparison.

\section{Sparse Distributed Representations}~\label{app:sdr}

The neocortex stores and represents information using sparse activity patterns, as demonstrated by empirical evidence~\cite{ahmad2015properties}. Inspired by HTM~\cite{hawkins2016neurons} and neuroscience-based theories of cortical function, we use Sparse Distributed Representations (SDRs) as the main representation format of PAM. An SDR is a sparse binary representation of a cell assembly where only a small fraction of the neurons in the SDR are active at any time. The location of these active neurons encodes the information that is represented by this SDR. In this section we describe some useful properties of SDRs and discuss their robustness to noise as opposed to dense representations.

\subsection{SDR Properties}

SDRs are used to represent rich sensory information in the neocortex as a sparse activity pattern. Therefore, from the mathematical viewpoint, an SDR must have the ability to represent many patterns and easily distinguish between them. The capacity of an SDR can be calculated as the possible combinations of locations where neurons can be active. Consider an SDR with size $N$ and number of active neurons $W$. The total capacity of this SDR is computed as shown in Equation~\ref{eqn:sdr_capacity}.

\begin{equation}\label{eqn:sdr_capacity}
    {N \choose W} = \frac{N!}{W!(N-W)!}
\end{equation}

Based on the above Binomial coefficient equation, it may seem that sparsity is not optimal for capacity as the capacity will be the highest when $W$ is exactly half of $N$. While capacity is important, we aim to represent multiple possibilities as a union of SDRs and therefore minimize the overlap between them. From an information-theoretic viewpoint, the goal is to minimize \textit{mutual information} between SDRs to ensure that each SDR carries unique information and the union represents a more comprehensive and diverse set of features. We can minimize the expected IoU by using lower sparsities as shown in Theorem~\ref{theorem:iou}. In our experiments we use $N=100$ and $W=5$, which results in capacity of $\approx \num{75e6}$ and an expected IoU of $\approx 0.02$. However, when scaled up to more typical values of SDR sizes and sparsities in the neocortex~\cite{hawkins2016neurons, ahmad2015properties} (i.e., $N=2048$, $W=40$), we get capacity of $\approx \num{2.37e84}$ (more than the estimated number of atoms in the observable universe $\approx 10^{80}$) and expected IoU of $\approx 0.01$. A sparsity of $0.5$ maximizes the mutual information and results in an expected IoU of $0.33$ which cannot be used to represent multiple possibilities as a union of SDRs, in spite of the optimal capacity. In practice, the size $N$ of the SDR is increased to increase the capacity, and the sparsity $W/N$ is decreased to minimize the expected overlap.

\subsection{The robustness of SDRs}

\begin{figure}[H]
\centering
\includegraphics[width = \linewidth]{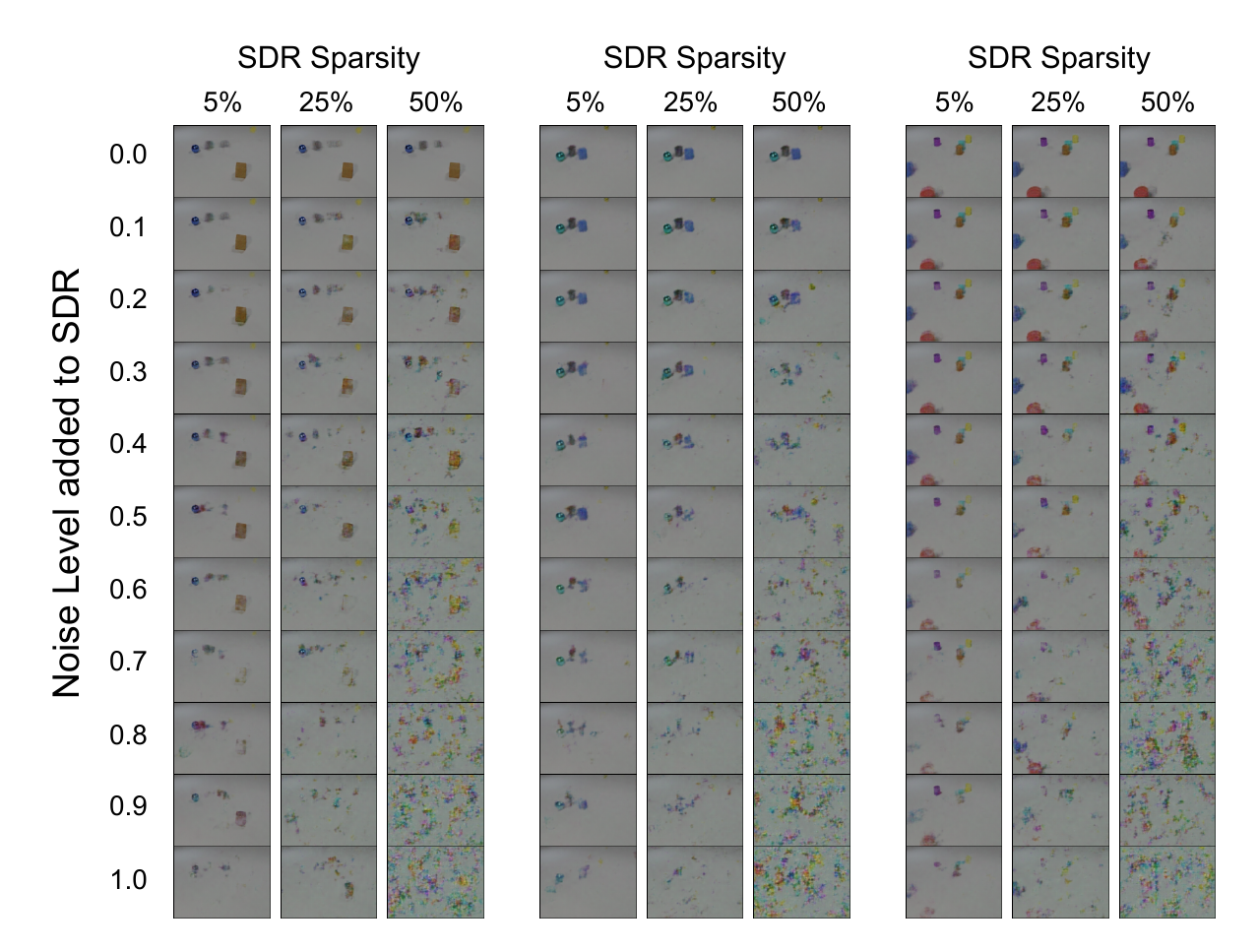}
\caption{Three examples of decoding an SDR with different noise levels. The results are shown for SDRs with different sparsities trained in an SDR autoencoder on CLEVRER dataset.} \label{fig:app_sdr_robustness}
\end{figure}

Sparse representations naturally minimize the overlap between random SDRs, therefore they are very tolerant to noise. To visualize this robustness property of SDRs, we design an experiment (Figure~\ref{fig:app_sdr_robustness}) where we train an SDR autoencoder with different sparsities and then decode SDRs at various levels of noise added. When the sparsity is increased to $50\%$, there is a high chance of overlap between SDRs, therefore a small amount of noise can cause collisions between SDRs. However, a $5\%$ sparsity can tolerate much more noise without overlapping with other SDRs.

\end{document}